\documentclass[lettersize,journal]{IEEEtran}
\usepackage{graphicx}
\usepackage{amsmath,amsfonts}
\usepackage{amssymb}
\usepackage{algorithm}
\usepackage{algorithmic}
\usepackage{hyperref}
\usepackage{array}
\usepackage[caption=false,font=normalsize,labelfont=sf,textfont=sf]{subfig}
\usepackage{textcomp}
\usepackage{stfloats}
\usepackage{url}
\usepackage{verbatim}
\usepackage{graphicx}
\usepackage[sorting=none, maxbibnames=3]{biblatex}
\usepackage{bbm}
\usepackage{soul}
\usepackage{xcolor}
\usepackage{pifont}

\usepackage{xcolor,colortbl}
\usepackage{hyperref}
\hypersetup{
     colorlinks=true,   
     linkcolor=blue,
     filecolor=magenta,      
     urlcolor=blue,
     citecolor=blue,
 }
\usepackage{balance}

\usepackage{balance}
\newcommand{\xmark}{\ding{55}}%

\newcommand\myfunc[5]{%
  \begingroup
  \setlength\arraycolsep{0pt}
  #1\colon\begin{array}[t]{c >{{}}c<{{}} l}
             #2 & \to & #3 \\ #4 & \mapsto & #5 
          \end{array}%
  \endgroup}

\title{Advances and Challenges in Meta-Learning: A Technical Review}
\author{
\IEEEauthorblockN{
Anna Vettoruzzo\IEEEauthorrefmark{1}, 
Mohamed-Rafik Bouguelia\IEEEauthorrefmark{1}, 
Joaquin Vanschoren\IEEEauthorrefmark{2}, 
Thorsteinn Rögnvaldsson\IEEEauthorrefmark{1},
KC Santosh\IEEEauthorrefmark{3}}\\
\IEEEauthorblockA{\IEEEauthorrefmark{1}\textit{Center for Applied Intelligent Systems Research (CAISR), Halmstad University, Sweden} \\\{anna.vettoruzzo, mohamed-rafik.bouguelia, thorsteinn.rognvaldsson\}@hh.se} \\
\IEEEauthorblockA{\IEEEauthorrefmark{2}\textit{Automated Machine Learning Group, Eindhoven University of Technology, Netherlands} \\\ j.vanschoren@tue.nl} \\
\IEEEauthorblockA{\IEEEauthorrefmark{3}\textit{Applied AI Research Lab, Department of Computer Science, University of South Dakota, USA} \\\ santosh.kc@usd.edu}
}

\begin{document}

\maketitle

\begin{abstract}
Meta-learning empowers learning systems with the ability to acquire knowledge from multiple tasks, enabling faster adaptation and generalization to new tasks. This review provides a comprehensive technical overview of meta-learning, emphasizing its importance in real-world applications where data may be scarce or expensive to obtain. The paper covers the state-of-the-art meta-learning approaches and explores the relationship between meta-learning and multi-task learning, transfer learning, domain adaptation and generalization, self-supervised learning, personalized federated learning, and continual learning. By highlighting the synergies between these topics and the field of meta-learning, the paper demonstrates how advancements in one area can benefit the field as a whole, while avoiding unnecessary duplication of efforts. Additionally, the paper delves into advanced meta-learning topics such as learning from complex multi-modal task distributions, unsupervised meta-learning, learning to efficiently adapt to data distribution shifts, and continual meta-learning. 
Lastly, the paper highlights open problems and challenges for future research in the field. By synthesizing the latest research developments, this paper provides a thorough understanding of meta-learning and its potential impact on various machine learning applications. We believe that this technical overview will contribute to the advancement of meta-learning and its practical implications in addressing real-world problems.
\end{abstract}


\section{Introduction}
\subsection*{Context and motivation}

Deep representation learning has revolutionized the field of machine learning by enabling models to learn effective features from data. However, it often requires large amounts of data for solving a specific task, making it impractical in scenarios where data is scarce or costly to obtain. Most existing approaches rely on either supervised learning of a representation tailored to a single task, or unsupervised learning of a representation that captures general features that may not be well-suited to new tasks. Furthermore, learning from scratch for each task is often not feasible, especially in domains such as medicine, robotics, and rare language translation where data availability is limited.

To overcome these challenges, meta-learning has emerged as a promising approach. Meta-learning enables models to quickly adapt to new tasks, even with few examples, and generalize across them. While meta-learning shares similarities with transfer learning and multitask learning, it goes beyond these approaches by enabling a learning system to \emph{learn how to learn}. This capability is particularly valuable in settings where data is scarce, costly to obtain, or where the environment is constantly changing. While humans can rapidly acquire new skills by leveraging prior experience and are therefore considered \emph{generalists}, most deep learning models are still \emph{specialists} and are limited to performing well on specific tasks. Meta-learning bridges this gap by enabling models to efficiently adapt to new tasks.

\subsection*{Contribution}
This review paper primarily discusses the use of meta-learning techniques in deep neural networks to learn reusable representations, with an emphasis on few-shot learning; it does not cover topics such as AutoML and Neural Architecture Search \cite{karmaker2021automl}, which are out of scope. 
Similarly, even though meta-learning is often applied in the context of reinforcement learning \cite{finn2017model, beck2023survey}, it falls outside the scope of this paper.
Distinct from existing surveys on meta-learning, such as \cite{vanschoren2018meta, hospedales2021meta, vilalta2002perspective, huisman2021survey, ZOU20231}, this review paper highlights several key differentiating factors:
\begin{itemize}
\item \textbf{Inclusion of advanced meta-learning topics}. In addition to covering fundamental aspects of meta-learning, this review paper delves into advanced topics such as learning from multimodal task distributions, meta-learning without explicit task information, learning without data sharing among clients, adapting to distribution shifts, and continual learning from a stream of tasks. By including these advanced topics, our paper provides a comprehensive understanding of the current state-of-the-art and highlights the challenges and opportunities in these areas.
\item \textbf{Detailed exploration of relationship with other topics}. We not only examine meta-learning techniques but also establish clear connections between meta-learning and related areas, including transfer learning, multitask learning, self-supervised learning, personalized federated learning, and continual learning. This exploration of the relationships and synergies between meta-learning and these important topics provides valuable insights into how meta-learning can be efficiently integrated into broader machine learning frameworks.
\item \textbf{Clear and concise exposition}. Recognizing the complexity of meta-learning, this review paper provides a clear and concise explanation of the concepts, techniques and applications of meta-learning. 
It is written with the intention of being accessible to a wide range of readers, including both researchers and practitioners. Through intuitive explanations, illustrative examples, and references to seminal works, we facilitate readers' understanding of the foundation of meta-learning and its practical implications.
\item \textbf{Consolidation of key information}. 
As a fast-growing field, meta-learning has information scattered across various sources. 
This review paper consolidates the most important and relevant information about meta-learning, presenting a comprehensive overview in a single resource. By synthesizing the latest research developments, this survey becomes an indispensable guide to researchers and practitioners seeking a thorough understanding of meta-learning and its potential impact on various machine learning applications.
\end{itemize}
By highlighting these contributions, this paper complements existing surveys and offers unique insights into the current state and future directions of meta-learning.

\subsection*{Organization}
In this paper, we provide the foundations of modern deep learning methods for learning across tasks. To do so, we first define the key concepts and introduce relevant notations used throughout the paper in section \ref{sec:notations}. Then, we cover the basics of multitask learning and transfer learning and their relation to meta-learning in section \ref{sec:mtl-ml}. In section \ref{sec:mt-approaches}, we present an overview of the current state of meta-learning methods and provide a unified view that allows us to categorize them into three types: black-box meta-learning methods, optimization-based meta-learning methods, and meta-learning methods that are based on distance metric learning \cite{vinalys}. In section \ref{sec:advanced-topics}, we delve into advanced meta-learning topics, explaining the relationship between meta-learning and other important machine learning topics, and addressing issues such as learning from multimodal task distributions, performing meta-learning without provided tasks, learning without sharing data across clients, learning to adapt to distribution shifts, and continual learning from a stream of tasks. Finally, the paper explores the application of meta-learning to real-world problems and provides an overview of the landscape of promising frontiers and yet-to-be-conquered challenges that lie ahead. Section \ref{sec:open-chal} focuses on these challenges, shedding light on the most pressing questions and future research opportunities.

\section{Basic notations and definitions} \label{sec:notations}

In this section, we introduce some simple notations which will be used throughout the paper and provide a formal definition of the term ``task" within the scope of this paper.

We use $\theta$ (and sometimes also $\phi$) to represent the set of parameters (weights) of a deep neural network model. $\mathcal{D} = \{ (x_j, y_j) \}_{j=1}^n$ denotes a dataset, where inputs $x_j$ are sampled from the distribution $p(x)$ and outputs $y_j$ are sampled from $p(y|x)$. The function $\mathcal{L}(., .)$ denotes a loss function, for example, $\mathcal{L}(\theta, \mathcal{D})$ represents the loss achieved by the model's parameters $\theta$ on the dataset $\mathcal{D}$. The symbol $\mathcal{T}$ refers to a task, which is primarily defined by the data-generating distributions $p(x)$ and $p(y|x)$ that define the problem.

In a standard supervised learning scenario, the objective is to optimize the parameters $\theta$ by minimizing the loss $\mathcal{L}(\theta, \mathcal{D})$, where the dataset $\mathcal{D}$ is derived from a single task $\mathcal{T}$, and the loss function $\mathcal{L}$ depends on that task. Formally, in this setting, a task $\mathcal{T}_i$ is a triplet $\mathcal{T}_i \triangleq \{ p_i(x), p_i(y|x), \mathcal{L}_i \}$ that includes task-specific data-generating distributions $p_i(x)$ and $p_i(y|x)$, as well as a task-specific loss function $\mathcal{L}_i$. The goal is to learn a model that performs well on data sampled from task $\mathcal{T}_i$. In a more challenging setting, we consider learning from multiple tasks $\{ \mathcal{T}_i \}_{i=1}^T$, which involves (a dataset of) multiple datasets $\{ \mathcal{D}_i \}_{i=1}^T$. 
In this scenario, a set of training tasks is used to learn a model that performs well on test tasks. Depending on the specific setting, a test task can either be sampled from the training tasks or completely new, never encountered during the training phase.

In general, tasks can differ in various ways depending on the application. For example, in image recognition, different tasks can involve recognizing handwritten digits or alphabets from different languages \cite{snell2017prototypical, finn2017model}, while in natural language processing, tasks can include sentiment analysis \cite{geng2019induction, liang2023few}, machine translation \cite{gu2020meta}, and chatbot response generation \cite{madotto2019personalizing, qian2019domain, mi2019meta}. Tasks in robotics can involve training robots to achieve different goals \cite{finn2017one}, while in automated feedback generation, tasks can include providing feedback to students on different exams \cite{wu2021prototransformer}. It is worth noting that tasks can share structures, even if they appear unrelated. For example, the laws of physics underlying real data, the language rules underlying text data, and the intentions of people all share common structures that enable models to transfer knowledge across seemingly unrelated tasks.

\section{From multitask and transfer to meta-learning} \label{sec:mtl-ml}
Meta-learning, multitask learning, and transfer learning encompass different approaches aimed at learning across multiple tasks. Multitask learning aims to improve performance on a set of tasks by learning them simultaneously. Transfer learning fine-tunes a pre-trained model on a new task with limited data. In contrast, meta-learning acquires useful knowledge from past tasks and leverages it to learn new tasks more efficiently. In this section, we transition from discussing ``multitask learning" and ``transfer learning" to introducing the topic of ``meta-learning".

\begin{figure}[tbp]
\centering
\includegraphics[width=\columnwidth]{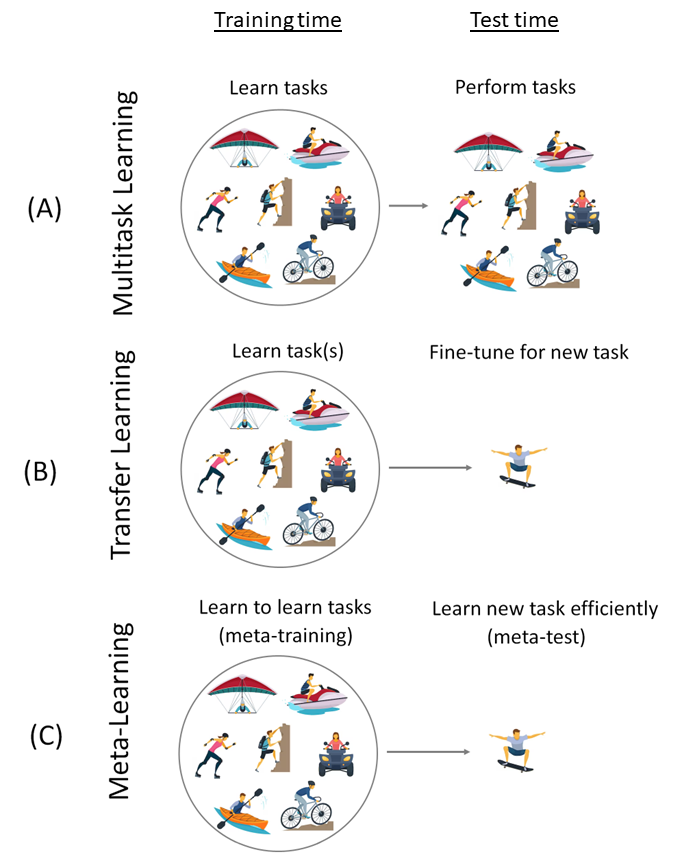}
\caption{Multitask learning vs transfer learning vs meta-learning} \label{fig:illustration-mtl-tl-ml}
\end{figure}

\subsection{Multitask learning problem} \label{sec:mtl}

As illustrated in Figure \ref{fig:illustration-mtl-tl-ml} (A), multitask learning (MTL) trains a model to perform multiple related tasks simultaneously, leveraging shared structure across tasks, and improving performance compared to learning each task individually. In this setting, there is no distinction between training and test tasks, and we refer to them as $\{\mathcal{T}_i\}_{i=1}^T$.

One common approach in MTL is hard parameter sharing, where the model parameters $\theta$ are split into shared $\theta^{\text{sh}}$ and task-specific $\theta^i$ parameters. These parameters are learned simultaneously through an objective function that takes the form: $$\min_{\theta^{\text{sh}}, \theta^1, \dots, \theta^T} \sum_{i=1}^{T} w_i \mathcal{L}_i(\{ \theta^{\text{sh}}, \theta^i \}, \mathcal{D}_i),$$ where $w_i$ can weight tasks differently. This approach is often implemented using a multi-headed neural network architecture, where a shared encoder (parameterized by $\theta^{\text{sh}}$) is responsible for feature extraction. This shared encoder subsequently branches out into task-specific decoding heads (parameterized by $\theta^i$) dedicated to individual tasks $\mathcal{T}_i$ \cite{sener2018multi, chen2018gradnorm, kendall2018multi}.

Soft parameter sharing is another approach in MTL that encourages parameter similarity across task-specific models using regularization penalties \cite{misra2016cross,ruder2019latent,gao2019nddr}. In this approach, each task typically has its own model with its own set of parameters $\theta^i$, while the shared parameters set $\theta^{\text{sh}}$ can be empty. The objective function is similar to that of hard parameter sharing, but with an additional regularization term that controls the strength of parameter sharing across tasks. The strength of regularization is determined by the hyperparameter $\lambda$. In the case of $L_2$ regularization, the objective function is given by: $$\min_{\theta^{\text{sh}}, \theta^1, \dots, \theta^T} \sum_{i=1}^{T} w_i \mathcal{L}_i(\{ \theta^{\text{sh}}, \theta^i \}, \mathcal{D}_i) + \lambda \sum_{i'=1}^T \lVert \theta^i - \theta^{i'} \rVert.$$ However, soft parameter sharing can be more memory-intensive as separate sets of parameters are stored for each task, and it requires additional design decisions and hyperparameters.

Another approach to sharing parameters is to condition a single model on a task descriptor $z_i$ that contains task-specific information used to modulate the network's computation. The task descriptor $z_i$ can be a simple one-hot encoding of the task index or a more complex task specification, such as language description or user attributes. When a task descriptor is provided, it is used to modulate the weights of the shared network with respect to the task at hand. Through this modulation mechanism, the significance of the shared features is determined based on the particular task, enabling the learning of both shared and task-specific features in a flexible manner. Such an approach grants fine-grained control over the adjustment of the network's representation, tailoring it to each individual task.
Various methods for conditioning the model on the task descriptor are described in \cite{dumoulin2018feature-wise}. More complex methods are also provided in \cite{liu2019end, long2017learning, jaegleperceiver}. 

Choosing the appropriate approach for parameter sharing, determining the level of the network architecture at which to share parameters, and deciding on the degree of parameter sharing across tasks are all design decisions that depend on the problem at hand. Currently, these decisions rely on intuition and knowledge of the problem, making them more of an art than a science, similar to the process of tuning neural network architectures. Moreover, multitask learning presents several challenges, such as determining which tasks are complementary, particularly in scenarios with a large number of tasks, as in \cite{fifty2021efficiently}. Interested readers can find a more comprehensive discussion of multitask learning in \cite{zhang2021survey, crawshaw2020multi}. 

In summary, multitask learning aims to learn a set of $T$ tasks $\{ \mathcal{T}_i \}_{i=1}^T$ at once. Even though the model can generalize to new data from these $T$ tasks, it might not be able to handle a completely new task that it has not been trained on. This is where transfer learning and meta-learning become more relevant. 

\subsection{Transfer learning via fine-tuning} \label{sec:tlvft}

Transfer learning is a valuable technique that allows a model to leverage representations learned from one or more source tasks to solve a target task. As illustrated in Figure \ref{fig:illustration-mtl-tl-ml} (B), the main goal is to use the knowledge learned from the source task(s) $\mathcal{T}_a$ to improve the performance of the model on a new task, usually referred to as the target task $\mathcal{T}_b$, especially when the target task dataset $\mathcal{D}_b$ is limited. In practice, the source task data $\mathcal{D}_a$ is often inaccessible, either because it is too expensive to obtain or too large to store. 

One common approach for transfer learning is fine-tuning, which involves starting with a model that has been pre-trained on the source task dataset $\mathcal{D}_a$. The parameters of the pre-trained model, denoted as $\theta$, are then fine-tuned on the training data $\mathcal{D}_b$ from the target task $\mathcal{T}_b$ using gradient descent or any other optimizer for several optimization steps. An example of the fine-tuning process for one gradient descent step is expressed as follows:
$$\phi \leftarrow \theta - \alpha \nabla_\theta \mathcal{L}(\theta, \mathcal{D}_b),$$
where $\phi$ denotes the parameters fine-tuned for task $\mathcal{T}_b$, and $\alpha$ is the learning rate.

Models with pre-trained parameters $\theta$ are often available online, including models pre-trained on large datasets such as ImageNet for image classification \cite{huh2016makes} and language models like BERT \cite{devlin2018bert}, PaLM \cite{chowdhery2022palm}, LLaMA \cite{touvron2023llama}, and GPT-4 \cite{OpenAI2023GPT4TR}, trained on large text corpora. Models pre-trained on other large and diverse datasets or using unsupervised learning techniques, as discussed in section \ref{unsup-meta-learn}, can also be used as a starting point for fine-tuning.

However, as discussed in \cite{kumarfine}, it is crucial to avoid destroying initialized features when fine-tuning. Some design choices, such as using a smaller learning rate for earlier layers, freezing earlier layers and gradually unfreezing, or re-initializing the last layer, can help to prevent this issue. Recent studies such as \cite{lee2022surgical} show that fine-tuning the first or middle layers can sometimes work better than fine-tuning the last layers, while others recommend a two-step process of training the last layer first and then fine-tuning the entire network \cite{kumarfine}. More advanced approaches, such as STILTs \cite{phang2018sentence}, propose an intermediate step of further training the model on a labeled task with abundant data to mitigate the potential degradation of pre-trained features.

In \cite{howard2018universal}, it was demonstrated that transfer learning via fine-tuning may not always be effective, particularly when the target task dataset is very small or very different from the source tasks. To investigate this, the authors fine-tuned a pre-trained universal language model on specific text corpora corresponding to new tasks using varying numbers of training examples. Their results showed that starting with a pre-trained model outperformed training from scratch on the new task. However, when the size of the new task dataset was very small, fine-tuning on such a limited number of examples led to poor generalization performance. To address this issue, meta-learning can be used to learn a model that can effectively adapt to new tasks with limited data by leveraging prior knowledge from other tasks. In fact, meta-learning is particularly useful for learning new tasks from very few examples, and we will discuss it in more detail in the remainder of this paper.

\subsection{Meta-learning problem}

Meta-learning (or learning to learn) is a field that aims to surpass the limitations of traditional transfer learning by adopting a more sophisticated approach that explicitly optimizes for transferability.
As discussed in section \ref{sec:tlvft}, traditional transfer learning involves pre-training a model on source tasks and fine-tuning it for a new task. In contrast, meta-learning trains a network to efficiently learn or adapt to new tasks with only a few examples. Figure \ref{fig:illustration-mtl-tl-ml} (C) illustrates this approach, where at meta-training time we \emph{learn to learn} tasks, and at meta-test time we \emph{learn} a new task efficiently.

During the meta-training phase, prior knowledge enabling efficient learning of new tasks is extracted from a set of training tasks $\{ \mathcal{T}_i \}_{i=1}^T$. This is achieved by using a meta-dataset consisting of multiple datasets $\{ \mathcal{D}_i \}_{i=1}^T$, each corresponding to a different training task. At meta-test time, a small training dataset $\mathcal{D}_{\text{new}}$ is observed from a completely new task $\mathcal{T}_{\text{new}}$ and used in conjunction with the prior knowledge to infer the most likely posterior parameters. As in transfer learning, accessing prior tasks at meta-test time is impractical. Although the datasets $\{ \mathcal{D}_i \}_i$ come from different data distributions (since they come from different tasks $\{ \mathcal{T}_i \}_i$), it is assumed that the tasks themselves (both for training and testing) are drawn i.i.d. from an underlying task distribution $p(\mathcal{T})$, implying some similarities in the task structure. This assumption ensures the effectiveness of meta-learning frameworks even when faced with limited labeled data. 
Moreover, the more tasks that are available for meta-training, the better the model can learn to adapt to new tasks, just as having more data improves performance in traditional machine learning. 

In the next section, we provide a more formal definition of meta-learning and various approaches to it.

\section{Meta-learning methods} \label{sec:mt-approaches}

To gain a unified understanding of the meta-learning problem, we can draw an analogy to the standard supervised learning setting. In the latter, the goal is to learn a set of parameters $\phi$ for a base model $h_\phi$ (e.g., a neural network parametrized by $\phi$), which maps input data $x \in \mathcal{X}$ to the corresponding output $y \in \mathcal{Y}$ as follows:
\begin{equation}
    \myfunc{ h_{\phi} }{ \mathcal{X} }{ \mathcal{Y} }{ x }{ y = h_\phi(x). }
\end{equation}
To accomplish this, a typically large training dataset $\mathcal{D} = \{ (x_j, y_j) \}_{j=1}^n$ specific to a particular task $\mathcal{T}$ is used to learn $\phi$.

In the meta-learning setting, the objective is to learn prior knowledge, which consists of a set of meta-parameters $\theta$, for a procedure $\mathcal{F}_{\theta}(\mathcal{D}_i^{\text{tr}}, x^{\text{ts}})$. This procedure uses $\theta$ to efficiently learn from (or adapt to) a small training dataset $\mathcal{D}_i^{\text{tr}} = \{ (x_k, y_k) \}_{k=1}^K$ from a task $\mathcal{T}_i$, and then make accurate predictions on unlabeled test data $x^{\text{ts}}$ from the same task $\mathcal{T}_i$. 
As we will see in the following sections, $\mathcal{F}_{\theta}$ is typically composed of two functions:
(1) a meta-learner $f_\theta(.)$ that produces task-specific parameters $\phi_i \in \Phi$ from $\mathcal{D}_i^{\text{tr}} \in \mathcal{X}^K$, and (2) a base model $h_{\phi_i}(.)$ that predicts outputs corresponding to the data in $x^{\text{ts}}$:
\begin{equation}
    \myfunc{ f_{\theta} }{ \mathcal{X}^K }{ \Phi }{ \mathcal{D}_i^{\text{tr}} }{ \phi_i = f_\theta(\mathcal{D}_i^{\text{tr}}), }
\quad \quad 
\myfunc{ h_{\phi_i} }{ \mathcal{X} }{ \mathcal{Y} }{ x }{ y = h_{\phi_i}(x). }
\end{equation}
Note that the process of obtaining task-specific parameters $\phi_i = f_\theta(\mathcal{D}_i^{\text{tr}})$ is often referred to as ``\emph{adaptation}" in the literature, as it adapts to the task $\mathcal{T}_i$ using a small amount of data while leveraging the prior knowledge summarized in $\theta$.
The objective of meta-training is to learn the set of meta-parameters $\theta$. This is accomplished by using a meta-dataset $\{ \mathcal{D}_i \}_{i=1}^T$, which consists of a dataset of datasets, where each dataset $\mathcal{D}_i = \{ (x_j, y_j) \}_{j=1}^n$ is specific to a task $\mathcal{T}_i$.

The unified view of meta-learning presented here is beneficial because it simplifies the meta-learning problem by reducing it to the design and optimization of $\mathcal{F}_{\theta}$. Moreover, it facilitates the categorization of the various meta-learning approaches into three categories: black-box meta-learning methods, optimization-based meta-learning methods, and distance metric-based meta-learning methods (as discussed in \cite{vinalys}). An overview of these categories is provided in the subsequent sections. 

\subsection{Black-box meta-learning methods} \label{sec:bbox-ml}

\begin{figure}[tbp]
\centering
\includegraphics[width=0.35\textwidth]{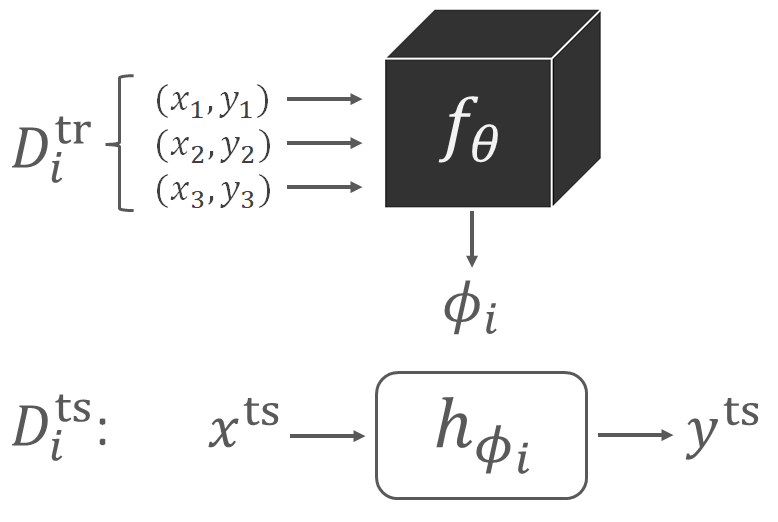}
\caption{Black-box meta-learning} \label{fig:black-box-ml}
\end{figure}

Black-box meta-learning methods, also known as model-based meta-learning \cite{huisman2021survey, vinalys}, represent $f_\theta$ as a black-box neural network that takes the entire training dataset, $\mathcal{D}_i^{\text{tr}}$, and predicts task-specific-parameters, $\phi_i$. These parameters are then used to parameterize the base network, $h_{\phi_i}$, and make predictions for test data-points, $y^{\text{ts}} = h_{\phi_i}(x^{\text{ts}})$. The architecture of this approach is shown in Figure \ref{fig:black-box-ml}. The meta-parameters, $\theta$, are optimized as shown in Equation \ref{eqn:bbox-optim}, and a general algorithm for these kinds of black-box methods is outlined in Algorithm \ref{alg:black-box-ml}.
\begin{equation} \label{eqn:bbox-optim}
\min_\theta \sum_{\mathcal{T}_i} \mathcal{L}( \underbrace{f_\theta(\mathcal{D}_i^{\text{tr}})}_{\phi_i}, \mathcal{D}_i^{\text{ts}} ).
\end{equation}

\begin{algorithm}[tbp]
\caption{Black-box meta-learning}\label{alg:black-box-ml}
\begin{algorithmic}[1]
\STATE Randomly initialize $\theta$
\WHILE {not done}
    \STATE Sample a task $\mathcal{T}_{i} \sim p(\mathcal{T})$ 
    ~~\textit{(or a mini-batch of tasks)}
    \STATE Sample disjoint datasets $\mathcal{D}_i^{\text{tr}}, \mathcal{D}_i^{\text{ts}}$ from $\mathcal{T}_{i}$
    \STATE Compute $\phi_i \gets f_\theta(\mathcal{D}_i^{\text{tr}})$
    \STATE Update $\theta$ using $\nabla_\theta \mathcal{L}(\phi_i, \mathcal{D}_i^{\text{ts}})$
\ENDWHILE
\STATE \textbf{return} $\theta$
\end{algorithmic}
\end{algorithm}

However, this approach faces a major challenge: outputting all the parameters $\phi_i$ of the base network $h_{\phi_i}$ is not scalable and is impractical for large-scale models. To overcome this issue, black-box meta-learning methods, such as MANN \cite{santoro2016meta} and SNAIL \cite{mishra2017simple}, only output sufficient statistics instead of the complete set of parameters of the base network. These methods allow $f_\theta$ to output a low-dimensional vector $z_i$ that encodes contextual task information, rather than a full set of parameters $\phi_i$. In this case, $\phi_i$ consists of $\{ z_i, \theta_h \}$, where $\theta_h$ denotes the trainable parameters of the network $h_{\phi_i}$. The base network $h_{\phi_i}$ is modulated with task descriptors by using various techniques for \emph{conditioning on task descriptors} discussed in section \ref{sec:mtl}.

Several black-box meta-learning methods adopt different neural network architectures to represent $f_\theta$. For instance, methods described in \cite{santoro2016meta}, use LSTMs or architectures with augmented memory capacities, such as Neural Turing Machines, while others, like Meta Networks \cite{munkhdalai2017meta}, employ external memory mechanisms. SNAIL \cite{mishra2017simple} defines meta-learner architectures that leverage temporal convolutions to aggregate information from past experience and attention mechanisms to pinpoint specific pieces of information. Alternatively, some methods, such as the one proposed in \cite{garnelo2018conditional}, use a feedforward plus averaging strategy. This latter feeds each data-point in $\mathcal{D}_i^{\text{tr}} = \{(x_j, y_j)\}_{j=1}^K$ through a neural network to produce a representation $r_j$ for each data-point, and then averages these representations to create a task representation $z_i = \frac{1}{K}\sum_{j=1}^K r_j$. This strategy may be more effective than using a recurrent model such as LSTM, as it does not rely on the assumption of temporal relationships between data-points in $\mathcal{D}_i^{\text{tr}}$.

Recent research efforts \cite{brown2020language, garg2022can, kirsch2022general, akyurek2022learning} have explored the connection between in-context learning and black-box meta-learning methods. This connection reveals that in-context learning can be viewed as a special instance of the broader meta-learning paradigm \cite{garg2022can}. In particular, in-context learning involves training models to perform well in new tasks with minimal examples, achieved by conditioning their response on context. Kirsch et al. \cite{kirsch2022general} demonstrate that general-purpose in-context learning algorithms can be trained from scratch using black-box models with minimal inductive bias (such as transformers \cite{vaswani2017attention}), highlighting the adaptability and potential of black-box meta-learning methods in these specialized contexts.

Black-box meta-learning methods are expressive, versatile, and easy to combine with various learning problems, including classification, regression, and reinforcement learning. However, they require complex architectures for the meta-learner $f_\theta$, making them computationally demanding and data-inefficient. As an alternative, one can represent $\phi_i = f_\theta(\mathcal{D}_i^{\text{tr}})$ as an optimization procedure instead of a neural network. The next section explores methods that utilize this approach.

\subsection{Optimization-based meta-learning methods} \label{sec:opt-ml}

\begin{figure}[tbp]
\centering
\includegraphics[width=\columnwidth]{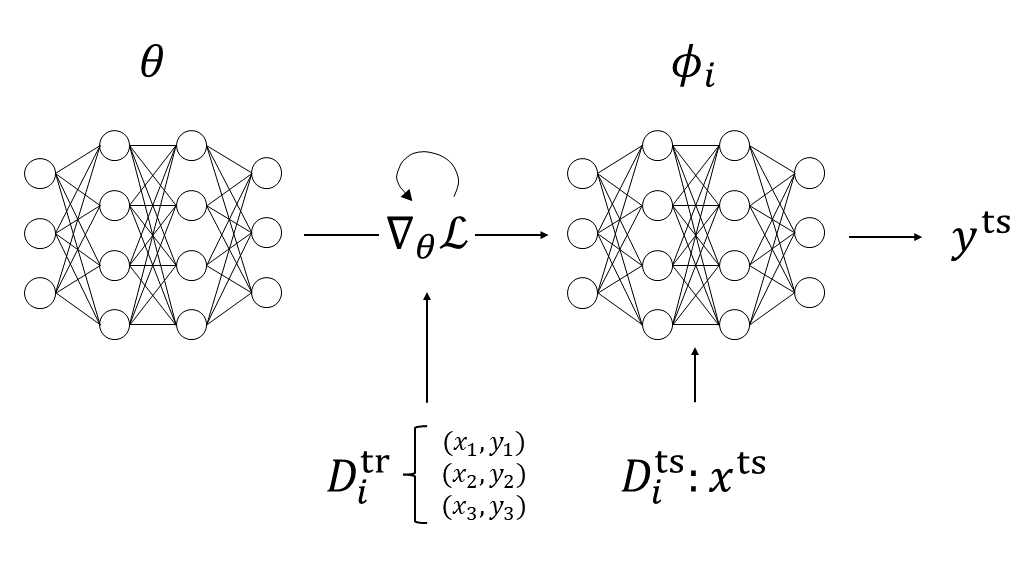}
\caption{Optimization-based meta-learning with gradient-based optimization} \label{fig:optimization-ml}
\end{figure}

Optimization-based meta-learning offers an alternative to the black-box approach, where the meta-learner $f_\theta$ is an optimization procedure like gradient descent, rather than a black-box neural network. The goal of optimization-based meta-learning is to acquire a set of meta-parameters $\theta$ that are easy to learn via gradient descent and to fine-tune on new tasks. Most optimization-based techniques do so by defining meta-learning as a bi-level optimization problem. At the inner level, $f_\theta$ produces task-specific parameters $\phi_i$ using $\mathcal{D}_i^{\text{tr}}$, while at the outer level, the initial set of meta-parameters $\theta$ is updated by optimizing the performance of $h_{{\phi}_i}$ on the test set of the same task. This is shown in Figure \ref{fig:optimization-ml} and in Algorithm \ref{alg:maml} in case $f_\theta$ is a gradient-based optimization. The meta-parameters $\theta$ can represent inner optimizers \cite{ravi2017optimization, andrychowicz2016learning, li2017learning, wichrowska2017learned}, neural network architectures \cite{shaw2019meta, lian2019towards}, other network hyperparameters \cite{franceschi2018bilevel}, or the initialization of the base model $h(.)$ \cite{finn2017model, kim2018bayesian}. The latter approach is similar to transfer learning via fine-tuning (cf. section \ref{sec:tlvft}), but instead of using a pre-trained $\theta$ that may not be transferable to new tasks, we learn $\theta$ to explicitly optimize for transferability.

Model-Agnostic Meta-Learning (MAML) \cite{finn2017model} is one of the earliest and most popular optimization-based meta-learning methods. The main idea behind MAML is to learn a set of initial neural network's parameters $\theta$ that can easily be fine-tuned for any task using gradient descent with only a few steps. During the meta-training phase, MAML minimizes the objective defined as follows:
\begin{equation} \label{eqn:maml}
\min_\theta \sum_{\mathcal{T}_i} \mathcal{L}( \underbrace{ \theta - \alpha \nabla_\theta \mathcal{L}(\theta, \mathcal{D}_i^{\text{tr}}) }_{\phi_i}, \mathcal{D}_i^{\text{ts}} ).
\end{equation}
Note that in Equation \ref{eqn:maml}, the task-specific parameters $\phi_i$ are obtained through a single gradient descent step from $\theta$, although in practice, a few more gradient steps are usually used for better performance.

As a result, MAML produces a model initialization $\theta$ that can be quickly adapted to new tasks with a small number of training examples. Algorithm \ref{alg:maml} can be viewed as a simplified illustration of MAML, where $\theta$ represents the parameters of a neural network. This is similar to Algorithm \ref{alg:black-box-ml} but with $\phi_i$ obtained through optimization.

During meta-test time, a small dataset $\mathcal{D}_{\text{new}}^{\text{tr}}$ is observed from a new task $\mathcal{T}_{\text{new}} \sim p(\mathcal{T})$. The goal is to use the prior knowledge encoded in $\theta$ to train a model that generalizes well to new, unseen examples from this task. To achieve this, $\theta$ is fine-tuned with a few adaptation steps using $\nabla_\theta \mathcal{L}(\theta, \mathcal{D}_{\text{new}}^{\text{tr}})$, resulting in task-specific parameters $\phi$. These parameters are then used to make accurate predictions on previously unseen input data from $\mathcal{T}_{\text{new}}$.

\begin{algorithm}[tbp]
\caption{Optimization-based meta-learning with gradient-based optimization}\label{alg:maml}
\begin{algorithmic}[1]
\STATE Randomly initialize $\theta$
\WHILE {not done}
    \STATE Sample a task $\mathcal{T}_{i} \sim p(\mathcal{T})$ 
    ~~\textit{(or a mini-batch of tasks)}
    \STATE Sample disjoint datasets $\mathcal{D}_i^{\text{tr}}, \mathcal{D}_i^{\text{ts}}$ from $\mathcal{T}_{i}$
    \STATE Optimize $\phi_i \gets \theta - \alpha \nabla_\theta \mathcal{L}(\theta, \mathcal{D}_i^{\text{tr}})$
    \STATE Update $\theta$ using $\nabla_\theta \mathcal{L}(\phi_i, \mathcal{D}_i^{\text{ts}})$
\ENDWHILE
\STATE \textbf{return} $\theta$
\end{algorithmic}
\end{algorithm}

\begin{figure}[tbp]
\centering
\includegraphics[width=0.45\textwidth]{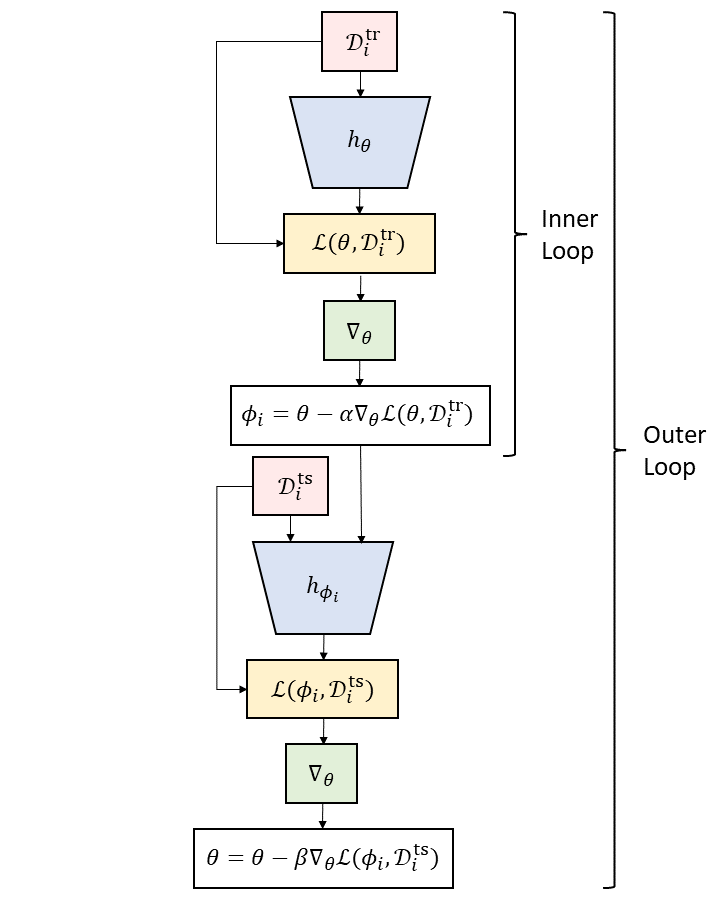}
\caption{Visual representation of the computation graph of MAML} \label{fig:computation_graph}
\end{figure}

MAML can be thought of as a computation graph (as shown in Figure \ref{fig:computation_graph}) with an embedded gradient operator. Interestingly, the components of this graph can be interchanged or replaced with components from the black-box approach. For instance, \cite{ravi2017optimization} also learned an initialization $\theta$, but adapted $\theta$ differently by using a learned network $f_w(\theta, \mathcal{D}_i^{\text{tr}}, \nabla_\theta \mathcal{L})$ instead of the gradient $\nabla_\theta \mathcal{L}(\theta, \mathcal{D}_i^{\text{tr}})$:
$$
\phi_i \gets \theta - \alpha f_w(\theta, \mathcal{D}_i^{\text{tr}}, \nabla_\theta \mathcal{L})
$$
In \cite{finn2017meta}, the authors investigated the effectiveness of optimization-based meta-learning in generalizing to similar but extrapolated tasks that are outside the original task distribution $p(\mathcal{T})$. The study found that, as task variability increases, black-box meta-learning methods such as SNAIL \cite{mishra2017simple} and MetaNet \cite{munkhdalai2017meta} acquire less generalizable learning strategies than gradient-based meta-learning approaches like MAML.

However, despite its success, MAML faces some challenges that have motivated the development of other optimization-based meta-learning methods. 
One of these challenges is the instability of MAML's bi-level optimization. Fortunately, there are enhancements that can significantly improve optimization process. For instance, Meta-SGD \cite{li2017meta} and AlphaMAML \cite{behl2019alpha} learn a vector of learning rates $\alpha$ automatically, rather than using a manually set scalar value $\alpha$. Other methods like DEML \cite{zhou2018deep}, ANIL \cite{raghu2019rapid} and BOIL \cite{oh2020boil} suggest optimizing only a subset of the parameters during adaptation. Additionally, MAML++ \cite{antoniou2018train} proposes various modifications to stabilize the optimization process and further improve the generalization performance. Moreover, Bias-transformation \cite{finn2017meta} and CAVIA \cite{zintgraf2019fast} introduce context variables for increased expressive power, while \cite{hiller2022enforcing} enforces a well-conditioned parameter space based on the concepts of the condition number \cite{goodfellow2016deep}.

Another significant challenge in MAML is the computationally expensive process of backpropagating through multiple gradient adaptation steps. To overcome this challenge, first-order alternatives to MAML such as FOMAML and Reptile have been introduced \cite{nichol2018first}. For example, Reptile aims to find an initialization $\theta$ that is close to each task's optimal parameters. Another approach is to optimize only the parameters of the last layer. For instance, \cite{bertinetto2018meta} and \cite{lee2019meta} perform a closed-form or convex optimization on top of meta-learned features. Another solution is iMAML \cite{rajeswaran2019meta}, which computes the full meta-gradient without differentiating through the optimization path, using the implicit function theorem.

\subsection{Meta-learning via distance metric learning} \label{sec:metric-ml}

In the context of low data regimes, such as in few-shot learning, simple non-parametric methods such as Nearest Neighbors \cite{Cover1967} can be effective. However, black-box and optimization-based meta-learning approaches discussed so far in sections \ref{sec:bbox-ml} and \ref{sec:opt-ml} have focused on using parametric base models, such as neural networks. In this section we discuss meta-learning approaches that employ a non-parametric learning procedure. The key concept is to use parametric meta-learners to produce effective non-parametric learners, thus eliminating the need for second-order optimization, as required by several methods discussed in section \ref{sec:opt-ml}.

\begin{figure}[tbp]
\centering
\includegraphics[width=0.9\columnwidth]{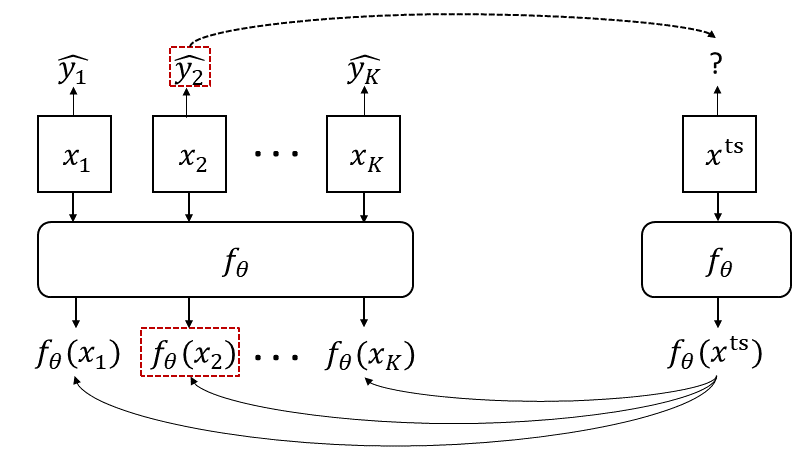}
\caption{Meta-learning via distance metric learning using matching network \cite{vinyals2016matching}} \label{fig:metric-ml}
\end{figure}

Suppose we are given a small training dataset $\mathcal{D}_i^{\text{tr}}$ that presents a 1-shot-$N$-way classification problem, i.e., $N$ classes with only one labeled data-point per class, along with a test data-point $x^{\text{ts}}$. To classify $x^{\text{ts}}$, a Nearest Neighbor learner compares it with each training data-point in $\mathcal{D}_i^{\text{tr}}$. However, determining an effective space and distance metric for this comparison can be challenging. For example, using the $L_2$ distance in pixel space for image data may not yield satisfactory results \cite{zhang2018unreasonable}. To overcome this, a distance metric can be derived by learning how to compare instances using meta-training data.
To learn an appropriate distance metric for comparing instances, a Siamese network \cite{koch2015siamese} can be trained to solve a binary classification problem that predicts whether two images belong to the same class. During meta-test time, each image in $\mathcal{D}_i^{\text{tr}}$ is compared with the test image $x^{\text{ts}}$ to determine whether they belong to the same class or not. However, there is a nuance due to the mismatch between the binary classification problem during meta-training and the $N$-way classification problem during meta-testing. Matching Networks, introduced in \cite{vinyals2016matching}, address this by learning an embedding space with a network $f_\theta$ and using Nearest Neighbors in the learned space, as shown in Figure \ref{fig:metric-ml}. The network is trained end-to-end to ensure that meta-training is consistent with meta-testing. Algorithm \ref{alg:matching-nets} outlines the meta-training process used by Matching Networks. It is similar to Algorithms \ref{alg:black-box-ml} and \ref{alg:maml}, except that the base model is non-parametric, so there is no $\phi_i$ (see lines \ref{alg:matching-nets:line:inner} and \ref{alg:matching-nets:line:outer}).

\begin{algorithm}[tbp]
\caption{Meta-learning via metric learning (matching networks)}\label{alg:matching-nets}
\begin{algorithmic}[1]
\STATE Randomly initialize $\theta$
\WHILE {not done}
    \STATE Sample a task $\mathcal{T}_{i} \sim p(\mathcal{T})$ 
    ~~\textit{(or a mini-batch of tasks)}
    \STATE Sample disjoint datasets $\mathcal{D}_i^{\text{tr}}, \mathcal{D}_i^{\text{ts}}$ from $\mathcal{T}_{i}$
    \STATE Compute $\hat{y}^{\text{ts}} = \sum\limits_{(x_k, y_k) \in \mathcal{D}_i^{\text{tr}}} f_\theta(x^{\text{ts}}, x_k) y_k$ \label{alg:matching-nets:line:inner}
    \STATE Update $\theta$ using $\nabla_\theta \mathcal{L}(\hat{y}^{\text{ts}}, y^{\text{ts}})$ \label{alg:matching-nets:line:outer}
\ENDWHILE
\STATE \textbf{return} $\theta$
\end{algorithmic}
\end{algorithm}

\begin{figure}[tbp]
\centering
\includegraphics[width=0.35\textwidth]{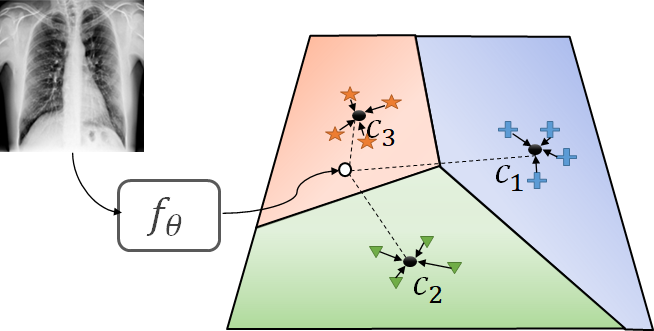}
\caption{Prototypical networks} \label{fig:proto-nets}
\end{figure}

However, Matching Networks are specifically designed for 1-shot classification and cannot be directly applied to $K$-shot classification problems (where there are $K$ labeled samples per class). To address this issue, other methods, such as Prototypical Networks \cite{laenen2021episodes}, have been proposed. Prototypical Networks aggregate class information to create a prototypical embedding, as illustrated in Figure \ref{fig:proto-nets}. In Prototypical Networks, line \ref{alg:matching-nets:line:inner} of Algorithm \ref{alg:matching-nets} is replaced with:
$$
p_\theta(y=l | x) = \frac{\exp{( -\lVert f_\theta(x) - c_l \rVert} )}{ \sum_{l'} \exp{( -\lVert f_\theta(x) - c_{l'} \rVert} ) },
$$
where $c_l$ is the mean embedding of all the samples in the $l$-th class, i.e., $c_l = \frac{1}{K} \sum_{(x, y) \in \mathcal{D}_i^{\text{tr}}} \mathbbm{1}(y=l) f_\theta(x)$.

While methods such as Siamese networks, Matching Networks, and Prototypical Networks can perform few-shot classification by embedding data and applying Nearest Neighbors \cite{koch2015siamese, vinyals2016matching, laenen2021episodes}, they may not be sufficient to capture complex relationships between data-points. To address this, alternative approaches have been proposed. RelationNet \cite{sung2018learning} introduces a non-linear relation module that can reason about complex relationships between embeddings. Garcia et al. \cite{garcia2017few} propose to use graph neural networks to perform message passing on embeddings, allowing for the capture of more complex dependencies. Finally, Allen et al. \cite{allen2019infinite} extend Prototypical Networks to learn an infinite mixture of prototypes, which improves the model's ability to represent the data distribution.

\subsection{Comparison and hybrid approaches}
\begin{table}
\begin{center}
\caption{Summary Comparison of Meta-Learning Approaches.}
\label{tab:summary}
\begin{tabular}{p{0.43\columnwidth} >{\centering\arraybackslash}p{0.123\columnwidth} >{\centering\arraybackslash}p{0.15\columnwidth} >{\centering\arraybackslash}p{0.12\columnwidth}}
\hline
& \textbf{BlackBox} & \textbf{Optimization} & \textbf{Metric}\\
\hline
Parametric Base Model & \textit{yes} & \textit{yes} & \textit{no} \\
Expressive Power & \textit{very high} & \textit{high} & \textit{high} \\
Consistency & \xmark & \checkmark & \checkmark \\
Versatility & \checkmark & \checkmark & \xmark \\
Simple Optimization & \xmark & \xmark & \checkmark \\
Data Efficiency & \xmark & \xmark & \checkmark \\
Positive Inductive Bias & \xmark & \checkmark & \xmark \\
Resource Efficiency & \xmark & \xmark & \checkmark \\
Generalizability to Tasks & \xmark & \checkmark & \xmark \\ 
\hline
\end{tabular}
\end{center}
\end{table}

Meta-learning approaches exhibit distinct strengths and tradeoffs. Black-box, optimization-based, and distance metric-based meta-learning approaches define $\mathcal{F}_{\theta}(\mathcal{D}_i^{\text{tr}}, x^{\text{ts}})$ differently, and the choice of method depends on specific use-case requirements. To assist readers, we provide a unified comparison in Table \ref{tab:summary}, summarizing the pros and cons of each approach based on the following criteria:

\begin{itemize}
    \item \textit{Parametric base model}: Indicates whether the approach is parametric (yes) or non-parametric (no), highlighting the model's flexibility.
    \item \textit{Expressive power}: Evaluates the ability of $\mathcal{F}_{\theta}$ to represent a wide range of learning procedures.
    \item \textit{Consistency}: Reflects the extent to which the learned learning procedure improves with additional data.
    \item \textit{Versatility}: Compatibility with different scenarios and the ease of integration with different learning problems, such as classification, regression, and reinforcement learning.
    \item \textit{Simple optimization}: Considers the complexity of optimization, including optimization challenges arising from complex models or the necessity for sophisticated second-order optimization techniques.
    \item \textit{Data efficiency (in terms of training tasks)}: The method's efficiency concerning the number of training tasks required for effective meta-learning.
    \item \textit{Positive inductive bias at meta-learning onset}: Indicates the presence of an inherent bias favoring certain learning strategies, influencing initial meta-learning performance.
    \item \textit{Resource efficiency}: Assesses computational and memory requirements, providing insights into the practical feasibility of the approach.
    \item \textit{Generalizability to diverse/variable tasks}: Explores the approach's ability to acquire generalizable learning strategies, extending to similar but extrapolated tasks beyond the original task distribution.
\end{itemize}

Although black-box, optimization-based, and distance metric-based meta-learning approaches differ, they are not mutually exclusive and can be combined in various ways.
For instance, in \cite{jiang2019learning}, gradient descent is applied while conditioning on the data, allowing the model to modulate the feature representations and capture inter-class dependencies. In \cite{rusu2018meta}, LEO (Latent Embedding Optimization) combines optimization-based meta-learning with a latent embedding produced by the RelationNet embedding proposed in \cite{sung2018learning}. The parameters of the model are first conditioned on the input data and then further adapted through gradient descent. In \cite{triantafillou2019meta}, the strength of both MAML and Prototypical Networks are combined to form a hybrid approach called Proto-MAML. This approach exploits the flexible adaptation of MAML, while initializing the last layer with ProtoNet to provide a simple inductive bias that is effective for very-few-shot learning. Similarly, \cite{wang2019hybrid} proposes a model where the meta-learner operates using an optimization-based meta-model, while the base learner exploits a metric-based approach (either Matching Network or Prototypical Network). The distance metrics used by the base learner can better adapt to different tasks thanks to the weight prediction from the meta-learner.

In summary, researchers have explored combining black-box, optimization-based, and distance metric-based meta-learning approaches to take advantage of their individual strengths. These combined approaches aim to improve performance, adaptability, and generalization in few-shot learning tasks by integrating different methodologies.

\section{Advanced meta-learning topics} \label{sec:advanced-topics}

The field of meta-learning has seen rapid development in recent years, with numerous methods proposed for learning to learn from a few examples. In this section, we delve into advanced topics in meta-learning that extend the meta-learning paradigm to more complex scenarios. We explore meta-learning from multi-modal task distributions, the challenge of out-of-distribution tasks, and unsupervised meta-learning. Additionally, we examine the relationship between meta-learning and personalized federated learning, domain adaptation/generalization, as well as the intersection between meta-learning and continual learning. By delving into these advanced topics, we can gain a deeper understanding of the potential of meta-learning and its applications in more complex real-world scenarios.

\subsection{Meta-learning from multimodal task distributions} \label{subsec:multimodal-distribution}

Meta-learning methods have traditionally focused on optimizing performance within a unimodal task distribution $p(\mathcal{T})$, assuming that all tasks are closely related and share similarities within a single application domain. 
However, in real-world scenarios, tasks are often diverse and sampled from a more complex task distribution with multiple unknown modes. The performance of most meta-learning approaches tends to deteriorate as the dissimilarity among tasks increases \cite{tian2020rethinking, Chen_2021_ICCV, guo2020broader, chen2019closer}, indicating that a globally shared set of meta-parameters $\theta$ may not adequately capture the heterogeneity among tasks and enable fast adaptation.

To address this challenge, MMAML \cite{vuorio2019multimodal} builds upon the standard MAML approach by estimating the mode of tasks sampled from a multimodal task distribution $p(\mathcal{T})$ and adjusting the initial model parameters accordingly. Another approach proposed in \cite{denevi2020advantage} involves learning a meta-regularization conditioned on additional task-specific information. However, obtaining such additional task information may not always be feasible.
Alternatively, some methods propose learning multiple model initializations ${\theta_1, \theta_2, \cdots, \theta_M}$ and selecting the most suitable one for each task, leveraging clustering techniques applied in either the task-space or parameter-space \cite{yao2019hierarchically, jiang2022subspace, jerfel2019reconciling, zhou2021task}, or relying on the output of an additional network, as in MUSE \cite{vettoruzzo2023meta}. CAVIA \cite{zintgraf2019fast} partitions the initial model parameters into shared parameters across all tasks and task-specific context parameters, while LGM-Net \cite{li2019lgm} directly generates classifier weights based on an encoded task representation.

A series of related works (but outside of the meta-learning field) aim to build a ``universal representation" that encompasses a robust set of features capable of achieving strong performance across multiple datasets (or modes) \cite{liu2020universal, li2021universal, requeima2019fast, garnelo2018conditional, dvornik2020selecting, dvornik2020selecting, triantafillou2021learning}. This representation is subsequently adapted to individual tasks in various ways. However, these approaches are currently limited to classification problems and do not leverage meta-learning techniques to efficiently adapt to new tasks. 

A more recent line of research focuses on cross-domain 
meta-learning, where knowledge needs to be transferred from tasks sampled from a potentially multimodal distribution $p(\mathcal{T})$ to target tasks sampled from a different distribution. One notable study, BOIL \cite{oh2020boil}, reveals that the success of meta-learning methods, such as MAML, can be attributed to large changes in the representation during task learning. The authors emphasize the importance of updating only the body (feature extractor) of the model and freezing the head (classifier) during the adaptation phase for effective cross-domain adaptation. Building on this insight, DAML \cite{lee2022domain} introduces tasks from both seen and pseudo-unseen domains during meta-training to obtain domain-agnostic initial parameters capable of adapting to novel classes in unseen domains. In \cite{kang2018transferable}, the authors propose a transferable meta-learning algorithm with a \emph{meta task adaptation} to minimize the domain divergence and thus facilitate knowledge transfer across domains. To further improve the transferability of cross-domain knowledge, \cite{li2021semi} and \cite{yuan2022task} propose to incorporate semi-supervised techniques into the meta-learning framework. Specifically, \cite{li2021semi} combines the representation power of large pre-trained language models (e.g., BERT \cite{devlin2018bert}) with the generalization capability of prototypical networks enhanced by SMLMT \cite{bansal2020self} to achieve effective generalization and adaptation to tasks from new domains. In contrast, \cite{yuan2022task} promotes the idea of task-level self-supervision by leveraging multiple views or augmentations of tasks.

\subsection{Meta-learning \& personalized federated learning}
Federated learning (FL) is a distributed learning paradigm where multiple clients collaborate to train a shared model while preserving data privacy by keeping their data locally stored. FedAvg \cite{mcmahan2017communication} is a pioneering method that combines local stochastic gradient descent on each client with model averaging on a central server. This approach performs well when local data across clients is independent and identically distributed (IID). However, in scenarios with heterogeneous (non-IID) data distributions, regularization techniques \cite{li2020federated, li2021model, karimireddy2020scaffold} have been proposed to improve local learning.

Personalized federated learning (PFL) is an alternative approach that aims to develop customized models for individual clients while leveraging the collaborative nature of FL. Popular PFL methods include L2GD \cite{hanzely2020federated}, which combines local and global models, as well as multi-task learning methods like pFedMe \cite{t2020personalized}, Ditto \cite{li2021ditto}, and FedPAC \cite{fedpac2023}. Clustered or group-based FL approaches \cite{ghosh2020efficient, duan2021flexible, sattler2020clustered, yang2023personalized} learn multiple group-based global models. In contrast, meta-learning-based methods interpret PFL as a meta-learning algorithm, where \emph{personalization to a client} aligns with \emph{adaptation to a task} \cite{jiang2019improving}. Notably, various combinations of MAML-type methods with FL architectures have been explored in \cite{fallah2020personalized, jiang2019improving, chen2018federated} to find an initial shared point that performs well after personalization to each client's local dataset. Additionally, the authors of \cite{khodak2019adaptive} proposed ARUBA, a meta-learning algorithm inspired by online convex optimization, which enhances the performance of FedAvg.

To summarize, there is a growing focus on addressing FL challenges in non-IID data settings. The integration of meta-learning has shown promising outcomes, leading to enhanced personalization and performance in PFL methods.

\subsection{Unsupervised meta-learning with tasks construction} \label{unsup-meta-learn}

\begin{figure}
\centering
\includegraphics[width=\columnwidth]{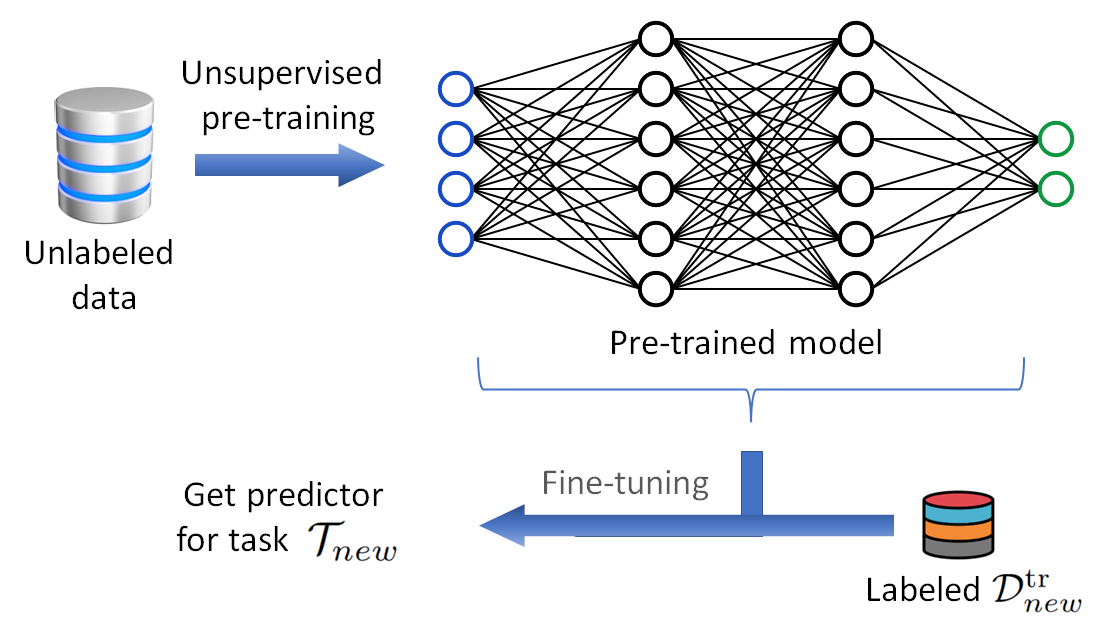}
\caption{Unsupervised pre-training} \label{fig:unsupervised-pre-training}
\end{figure}

In meta-training, constructing tasks typically relies on labeled data. However, real-world scenarios often involve mostly, or only, unlabeled data, requiring techniques that leverage unlabeled data to learn feature representations that can transfer to downstream tasks with limited labeled data. One alternative to address this is through ``self-supervised learning" (also known as ``unsupervised pre-training") \cite{chen2020simple, he2020momentum, grill2020bootstrap}. This involves training a model on a large unlabeled dataset, as depicted in Figure \ref{fig:unsupervised-pre-training}, to capture informative features. Contrastive learning \cite{oord2018representation, chen2020simple} is commonly used in this context, aiming to learn features by bringing similar examples closer together while pushing differing examples apart. The learned features can then be fine-tuned on a target task $\mathcal{T}_{\text{new}}$ with limited labeled data $\mathcal{D}_{\text{new}}^{\text{tr}}$, leading to improved performance compared to training from scratch. Another promising alternative is ``unsupervised meta-learning," which aims to automatically construct diverse and structured training tasks from unlabeled data. These tasks can then be used with any meta-learning algorithm, such as MAML \cite{finn2017model} and ProtoNet \cite{laenen2021episodes}. In this section, we explore methods for meta-training without predefined tasks and investigate strategies for automatically constructing tasks for meta-learning.

The method proposed in \cite{hsu2018unsupervised} constructs tasks based on unsupervised representation learning methods such as BiGAN \cite{donahue2016adversarial, donahue2019large} or DeepCluster \cite{caron2018deep} and clusters the data in the embedding space to assign pseudo-labels and construct tasks. Other methods such as UMTRA \cite{khodadadeh2019unsupervised} and LASIUM \cite{khodadadeh2020unsupervised} generate synthetic samples using image augmentations or pre-trained generative networks. In particular, the authors in \cite{khodadadeh2019unsupervised} construct a task $\mathcal{T}_i$ for a 1-shot $N$-way classification problem by creating a support set $\mathcal{D}_i^{\text{tr}}$ and a query set $\mathcal{D}_i^{\text{ts}}$ as follows:
\begin{itemize}
\item Randomly sample $N$ images and assign labels $1, \dots, N$, storing them in $\mathcal{D}_i^{\text{tr}}$.
\item Augment\footnote{Various augmentation techniques, like flipping, cropping, or reflecting an image, typically preserve its label. Likewise, nearby image patches or adjacent video frames share similar characteristics and are therefore assigned the same label.} each image in $\mathcal{D}_i^{\text{tr}}$, and store the resulting (augmented) images in $\mathcal{D}_i^{\text{ts}}$.
\end{itemize}
Such augmentations can be based on domain knowledge or learned augmentation strategies like those proposed in \cite{cubuk2019autoaugment}. 
In principle, task construction techniques can be applied beyond image-based augmentation. For instance, temporal aspects can be leveraged by incorporating time-contrastive learning on videos, as demonstrated in \cite{nair2022r3m}. Another approach is offered by Viewmaker Networks \cite{tamkin2020viewmaker}, which learn augmentations that yield favorable outcomes not only for images but also for speech and sensor data.
Contrary to these works focusing on generating pseudo tasks, Meta-GMVAE \cite{lee2021meta} and Meta-SVEBM \cite{kong2021unsupervised} address the problem by using variational autoencoders \cite{kingma2013auto} and energy-based models \cite{teh2003energy}, respectively. 
However, these methods are limited to the pseudo-labeling strategies used to create tasks, they rely on the quality of generated samples and they cannot scale to large-scale datasets.

To overcome this limitation, recent approaches have investigated the possibility of using self-supervised learning techniques to improve unsupervised meta-learning methods. In particular, in \cite{ni2021close}, the relationship between contrastive learning and meta-learning is explored, demonstrating that established meta-learning methods can achieve comparable performance to contrastive learning methods, and that representations transfer similarly well to downstream tasks. Inspired by these findings, the authors in \cite{yang2022few} integrate contrastive learning in a two-stage training paradigm consisting of sequential pre-training and meta-training stages. Another work \cite{leeself} interprets a meta-learning problem as a set-level problem and maximizes the agreement between augmented sets using SimCLR \cite{jang2023unsupervised}. Finally, PsCo \cite{jang2023unsupervised} builds upon MoCo \cite{he2020momentum} by progressively improving pseudo-labeling and constructing diverse tasks in an online manner. These findings indicate the potential for leveraging existing advances in meta-learning to improve contrastive learning (and vice-versa).

To meta-learn with unlabeled text data, some methods use language modeling, as shown in \cite{brown2020language} for GPT-3. Here, the support set $\mathcal{D}_i^{\text{tr}}$ consists of a sequence of characters, and the query set $\mathcal{D}_i^{\text{ts}}$ consists of the subsequent sequence of characters. However, this approach may not be suitable for text classification tasks, such as sentiment analysis or identifying political bias.
In \cite{bansal2020self}, an alternative approach (SMLMT) for self-supervised meta-learning for few-shot natural language classification tasks is proposed. SMLMT involves masking out words and classifying the masked word to construct tasks. The process involves: (1) sampling a subset of $N$ unique words and assigning each word a unique ID as its class label, (2) sampling $K+Q$ sentences that contain each of the $N$ words and masking out the corresponding word in each sentence, and (3) constructing the support set $\mathcal{D}_i^{\text{tr}}$ and the query set $\mathcal{D}_i^{\text{ts}}$ using the masked sentences and their corresponding word IDs.
SMLMT (for unsupervised meta-learning) is compared to BERT \cite{devlin2018bert}, a method that uses standard self-supervised learning and fine-tuning. SMLMT outperforms BERT on some tasks and achieves at least equal performance on others. Furthermore, Hybrid-SMLMT (semi-supervised meta-learning, which involves meta-learning on constructed tasks and supervised tasks), is compared to MT-BERT \cite{wu2021one} (multi-task learning on supervised tasks) and LEOPARD \cite{bansal2019learning} (an optimization-based meta-learner that uses only supervised tasks). The results show that Hybrid-SMLMT significantly outperforms these other methods. 

\subsection{Meta-learning \& domain adaptation/generalization} \label{ml-domain-adapt-gen}
Domain shift is a fundamental challenge, where the distribution of the input data changes between the training and test domains. To address this problem, there is a growing interest in utilizing meta-learning techniques for more effective domain adaptation and domain generalization. These approaches aim to enable models to quickly adapt to new domains with limited data or to train robust models that achieve better generalization on domains they have not been explicitly trained on. 
\smallskip

\subsubsection*{Effective domain adaptation via meta-learning}
Domain adaptation is a form of transductive transfer learning that leverages source domain(s) $p_S(x,y)$ to achieve high performance on test data from a target domain $p_T(x,y).$ It assumes $p_S(y|x) = p_T(y|x)$ but $p_S(x) \neq p_T(x)$, treating \emph{domains} as a particular kind of \emph{tasks}, with a task $\mathcal{T}_i \triangleq \{p_i(x), p_i(y|x), \mathcal{L}_i\}$ and a domain $d_i \triangleq \{p_i(x), p(y|x), \mathcal{L}\}$. For example, healthcare data from different hospitals with varying imaging techniques or patient demographics can correspond to different domains. Domain adaptation is most commonly achieved via feature alignment as in \cite{tzeng2014deep, ganin2016domain} or via translation between domains using CycleGAN \cite{zhu2017unpaired} as in \cite{rao2020rl, smith2019avid, hoffman2018cycada}. Other approaches focus on aligning the feature distribution of multiple source domains with the target domain \cite{zhao2018adversarial} or they address the multi-target domain adaptation scenario \cite{nguyen2021unsupervised, chen2019blending, gholami2020unsupervised} with models capable of adapting to multiple target domains. However, these methods face limitations when dealing with insufficient labeled data in the source domain or when quick adaptation to new target domains is required. Additionally, they assume the input-output relationship (i.e., $p(y|x)$) to be the same across domains. To solve these problems, some methods \cite{zhang2021adaptive, yang2022fewda, feng2021similarity, chen2019blending} combine meta-learning with domain adaptation. In particular, ARM \cite{zhang2021adaptive} leverages contextual information extracted from batches of unlabeled data to learn a model capable of adapting to distribution shifts. 
\smallskip

\subsubsection*{Effective domain generalization via meta-learning}

\begin{figure}[tbp]
\centering
\includegraphics[width=\columnwidth]{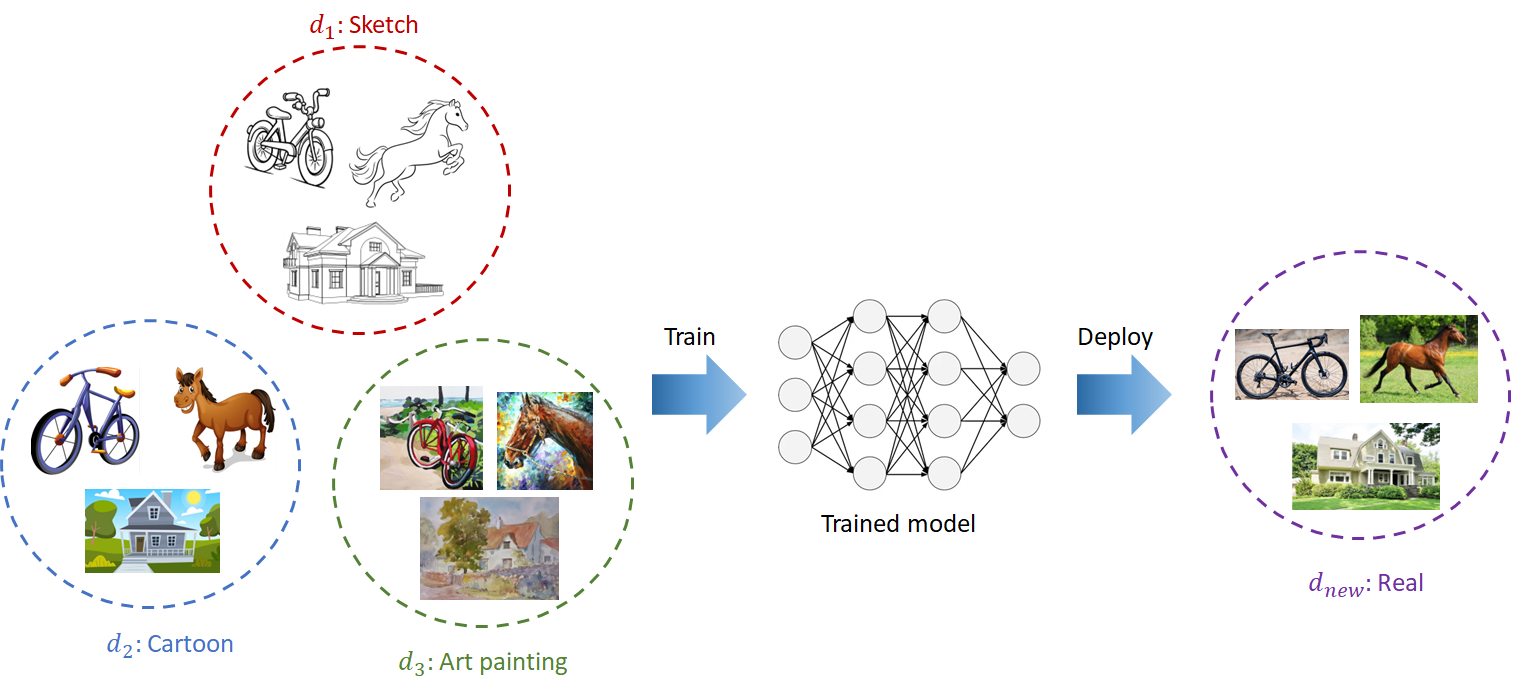}
\caption{Domain generalization problem} \label{fig:domain-generalization}
\end{figure}

Domain generalization enables models to perform well on new and unseen domains without requiring access to their data, as illustrated in Figure \ref{fig:domain-generalization}. This is particularly useful in scenarios where access to data is restricted due to real-time deployment requirements or privacy policies. For instance, an object detection model for self-driving cars trained on three types of roads may need to be deployed to a new road without any data from that domain. In contrast to domain adaptation, which requires access to (unlabeled) data from a specific target domain during training to specialize the model, domain generalization belongs to the inductive setting. Most domain generalization methods aim to train neural networks to learn domain-invariant representations that are consistent across domains. For instance, domain adversarial training \cite{sicilia2023domain} trains the network to make predictions based on features that cannot be distinguished between domains. Another approach is to directly align the representations between domains using similarity metrics, such as in \cite{sun2016deep}. Data augmentation techniques are also used to enhance the diversity of the training data and improve generalization across domains \cite{zhang2017mixup, verma2019manifold, yao2022improving}.
Another way to improve generalization to various domains is to use meta-learning and applying the episodic training paradigm typical of MAML \cite{finn2017meta}, as in \cite{dou2019domain, li2019episodic, li2019feature, li2018learning, balaji2018metareg, shu2021open, chen2022discriminative}. For instance, MLDG \cite{li2018learning} optimizes a model by simulating the train-test domain shift during the meta-training phase. MetaReg \cite{balaji2018metareg} proposes to meta-learn a regularization function that improves domain generalization. DADG \cite{chen2022discriminative} contains a discriminative adversarial learning component to learn a set of general features and a meta-learning-based cross-domain validation component to further enhance the robustness of the classifier.

\subsection{Meta-learning \& continual learning}

This section explores the application of meta-learning to continual learning, where learners continually accumulate experience over time to more rapidly acquire new knowledge or skills. Continual learning scenarios can be divided into task-incremental learning, domain-incremental learning, and class-incremental learning, depending on whether task identity is provided at test time or must be inferred by the algorithm \cite{van2022three}. In this section, we focus on approaches that specifically address task/class-incremental learning.

\begin{figure}[tbp]
\centering
\includegraphics[width=\columnwidth]{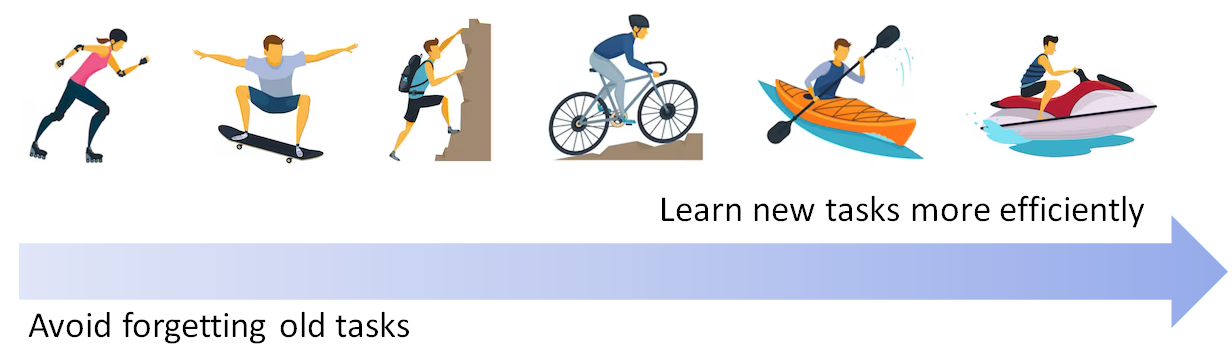}
\caption{Continual learning} \label{fig:sequential-tasks}
\end{figure}

Traditionally, meta-learning has primarily focused on scenarios where a batch of training tasks is available. However, real-world situations often involve tasks presented sequentially, allowing for progressive leveraging of past experience. This is illustrated in Figure \ref{fig:sequential-tasks}, and examples include tasks that progressively increase in difficulty or build upon previous knowledge, or robots learning diverse skills in changing environments.

Standard online learning involves observing tasks in a sequential manner, without any task-specific adaptation or use of past experience to accelerate adaptation. 
To tackle this issue, researchers have proposed various approaches, including memory-based methods \cite{lopez2017gradient, chaudhry2018efficient, rebuffi2017icarl}, regularization-based methods \cite{aljundi2018selfless, kirkpatrick2017overcoming, serra2018overcoming} and dynamic architectural methods \cite{li2019learn, pham2021contextual, rusu2016progressive}. However, each of these methods has its own limitations, such as scalability, memory inefficiency, time complexity, or the need for task-specific parameters. Meta-learning has emerged as a promising approach for addressing continual learning. In \cite{beaulieu2020learning}, the authors introduced ANML, a framework that meta-learns an activation-gating function that enables context-dependent selective activation within a deep neural network. This selective activation allows the model to focus on relevant knowledge and avoid catastrophic forgetting. Other approaches such as MER \cite{riemer2018learning}, OML \cite{javed2019meta}, and LA-MAML \cite{gupta2020look} use gradient-based meta-learning algorithms to optimize various objectives such as gradient alignment, inner representations, or task-specific learning rates and learn update rules that avoid negative transfer.
These algorithms enable faster learning over time and enhanced proficiency in each new task.

\section{Open challenges \& opportunities}\label{sec:open-chal}

Meta-learning has been a promising area of research that has shown impressive results in various machine learning domains. However, there are still open challenges that need to be addressed in order to further advance the field. In this section, we discuss some of these challenges and categorize them into three groups. Addressing these challenges can lead to significant advances in meta-learning, which could potentially lead to more generalizable and robust machine learning models.

\subsection{Addressing fundamental problem assumptions}
The first category of challenges pertains to the fundamental assumptions made in meta-learning problems. 

One such challenge is related to generalization to out-of-distribution tasks and long-tailed task distributions. Indeed, adaptation becomes difficult when the few-shot tasks observed at meta-test time are from a different task distribution than the ones seen during meta-training. While there have been some attempts to address this challenge, such as in \cite{prabhu2019few, kang2018transferable}, it still remains unclear how to address it. Ideas from the domain generalization and robustness literature could provide some hints and potentially be combined with meta-learning to tackle these long-tailed task distributions and out-of-distribution tasks. For example, possible directions are to define subtle regularization techniques to prevent the meta-parameters from being very specific to the distribution of the training tasks, or use subtle task augmentation techniques to generate synthetic tasks that cover a wider range of task variations.

Another challenge in this category involves dealing with the multimodality of data. While the focus has been on meta-training over tasks from a single modality, the reality is that we may have multiple modalities of data to work with. Human beings have the advantage of being able to draw upon multiple modalities, such as visual imagery, tactile feedback, language, and social cues, to create a rich repository of knowledge and make more informed decisions. For instance, we often use language cues to aid our visual decision-making processes. Rather than developing a prior that only works for a single modality, exploring the concept of learning priors across multiple modalities of data is a fascinating area to pursue. Different modalities have different dimensionalities or units, but they can provide complementary forms of information. While some initial works in this direction have been reported, including \cite{liang2021cross, alayrac2022flamingo, reed2022generalist}, there is still a long way to go in terms of capturing all of this rich prior information when learning new tasks.

\subsection{Providing benchmarks and real-world problems}
The second category of challenges is related to providing/improving benchmarks to better reflect real-world problems and challenges.

Meta-learning has shown promise in a diverse set of applications, including few-shot land cover classification \cite{russwurm2020meta}, few-shot dermatological disease diagnosis \cite{prabhu2019few}, automatically providing feedback on student code \cite{wu2021prototransformer}, one-shot imitation learning \cite{yu2018one}, drug discovery \cite{nguyen2020meta}, motion prediction \cite{gui2018few}, and language generation \cite{mi2019meta}, to mention but a few. However, the lack of benchmark datasets that accurately reflect real-world problems with appropriate levels of difficulty and ease of use is a significant challenge for the field. Several efforts have been made towards creating useful benchmark datasets, including Meta-Dataset \cite{triantafillou2019meta}, Meta-Album Dataset \cite{ullah2022meta}, NEVIS'22 \cite{bornschein2022nevis}, Meta-World Benchmark \cite{yu2020meta}, Visual Task Adaptation Benchmark \cite{zhai2019visual}, Taskonomy Dataset \cite{zamir2018taskonomy}, VALUE Benchmark \cite{li2021value}, and BIG Bench \cite{srivastava2022beyond}. However, further work is needed to ensure that the datasets are comprehensive and representative of the diversity of real-world problems that meta-learning aims to address.

Some ways with which existing benchmarks can be improved to better reflect real-world problems and challenges in meta-learning are: (1) to increase the diversity and complexity of tasks that are included; (2) to consider more realistic task distributions that can change over time; and (3) to include real-world data that is representative of the challenges faced in real-world applications of meta-learning. For example, including medical data, financial data, time-series data, or other challenging types of data (besides images and text) can help improve the realism and relevance of benchmarks. 

Furthermore, developing benchmarks that reflect these more realistic scenarios can help improve the generalization and robustness of algorithms. This ensures that algorithms are tested on a range of scenarios and that they are robust and generalizable across a wide range of tasks. Better benchmarks are essential for progress in machine learning and AI, as they challenge current algorithms to find common structures, reflect real-world problems, and have a significant impact in the real world.

\subsection{Improving core algorithms}
The last category of challenges in meta-learning is centered around improving the core algorithms.

A major obstacle is the large-scale bi-level optimization problem encountered in popular meta-learning methods such as MAML. The computational and memory costs of such approaches can be significant, and there is a need to make them more practical, particularly for very large-scale problems, like \emph{learning effective optimizers} \cite{metz2020tasks}.

In addition, a deeper theoretical understanding of various meta-learning methods and their performance is critical to driving progress and pushing the boundaries of the field. Such insights can inform and inspire further advancements in the field and lead to more effective and efficient algorithms. 
To achieve these goals, several fundamental questions can be explored, including:
\begin{enumerate}
\item Can we develop theoretical guarantees on the sample complexity and generalization performance of meta-learning algorithms? Understanding these aspects can help us design more efficient and effective meta-learning algorithms that require less data or less tasks. While recent investigations~\cite{chen2020closer, lucas2020theoretical, guan2022fast} have made notable strides in this domain, they represent just the initial steps toward a more extensive theoretical comprehension. Further research is imperative to completely harness the potential of meta-learning. 
\item Can we gain a better understanding of the optimization landscape of meta-learning algorithms? For instance, can we identify the properties of the objective function that make it easier or harder to optimize? Can we design optimization algorithms that are better suited to the bi-level optimization problem inherent in various meta-learning approaches?
\item Can we design meta-learning algorithms that can better incorporate task-specific or domain-specific expert knowledge, in a principled way, to learn more effective meta-parameters?
\end{enumerate}
Addressing such questions could enhance the design and performance of meta-learning algorithms, and help us tackle increasingly complex and challenging learning problems.

\section{Conclusion} \label{conclusion}
In conclusion, the field of artificial intelligence (AI) has witnessed significant advancements in developing specialized systems for specific tasks. However, the pursuit of generality and adaptability in AI across multiple tasks remains a fundamental challenge.

Meta-learning emerges as a promising research area that seeks to bridge this gap by enabling algorithms to learn how to learn. Meta-learning algorithms offer the ability to learn from limited data, transfer knowledge across tasks and domains, and rapidly adapt to new environments. This review paper has explored various meta-learning approaches that have demonstrated promising results in applications with scarce data. Nonetheless, numerous challenges and unanswered questions persist, calling for further investigation.

A key area of focus lies in unifying various fields such as meta-learning, self-supervised learning, domain generalization, and continual learning. Integrating and collaborating across these domains can generate synergistic advancements and foster a more comprehensive approach to developing AI systems. By leveraging insights and techniques from these different areas, we can construct more versatile and adaptive algorithms capable of learning from multiple tasks, generalizing across domains, and continuously accumulating knowledge.

This review paper serves as a starting point for encouraging research in this direction. By examining the current state of meta-learning and illuminating the challenges and opportunities, we aim to inspire researchers to explore interdisciplinary connections and contribute to the progress of meta-learning while integrating it with other AI research fields. Through collective efforts and collaboration, we can surmount existing challenges and unlock the full potential of meta-learning to address a broad spectrum of complex problems.

\section*{Acknowledgments}
This work was supported by the ``Knowledge Foundation'' (KK-stiftelsen).

\printbibliography

@article{dumoulin2018feature-wise,
  author = {Dumoulin, Vincent and Perez, Ethan and Schucher, Nathan and Strub, Florian and Vries, Harm de and Courville, Aaron and Bengio, Yoshua},
  title = {Feature-wise transformations},
  journal = {Distill},
  year = {2018},
  note = {https://distill.pub/2018/feature-wise-transformations},
}

@inproceedings{chen2018gradnorm,
  title={Gradnorm: Gradient normalization for adaptive loss balancing in deep multitask networks},
  author={Chen, Zhao and Badrinarayanan, Vijay and Lee, Chen-Yu and Rabinovich, Andrew},
  booktitle={International conference on machine learning},
  pages={794--803},
  year={2018},
  organization={PMLR}
}

@inproceedings{misra2016cross,
  title={Cross-stitch networks for multi-task learning},
  author={Misra, Ishan and Shrivastava, Abhinav and Gupta, Abhinav and Hebert, Martial},
  booktitle={Proceedings of the IEEE conference on computer vision and pattern recognition},
  pages={3994--4003},
  year={2016}
}

@inproceedings{liu2019end,
  title={End-to-end multi-task learning with attention},
  author={Liu, Shikun and Johns, Edward and Davison, Andrew J},
  booktitle={Proceedings of the IEEE/CVF conference on computer vision and pattern recognition},
  pages={1871--1880},
  year={2019}
}

@article{long2017learning,
  title={Learning multiple tasks with multilinear relationship networks},
  author={Long, Mingsheng and Cao, Zhangjie and Wang, Jianmin and Yu, Philip S},
  journal={Advances in neural information processing systems},
  volume={30},
  year={2017}
}

@inproceedings{jaegleperceiver,
  title={Perceiver IO: A General Architecture for Structured Inputs \& Outputs},
  author={Jaegle, Andrew and Borgeaud, Sebastian and Alayrac, Jean-Baptiste and Doersch, Carl and Ionescu, Catalin and Ding, David and Koppula, Skanda and Zoran, Daniel and Brock, Andrew and Shelhamer, Evan and others},
  booktitle={International Conference on Learning Representations}
}

@article{fifty2021efficiently,
  title={Efficiently identifying task groupings for multi-task learning},
  author={Fifty, Chris and Amid, Ehsan and Zhao, Zhe and Yu, Tianhe and Anil, Rohan and Finn, Chelsea},
  journal={Advances in Neural Information Processing Systems},
  volume={34},
  pages={27503--27516},
  year={2021}
}

@article{zhang2021survey,
  title={A survey on multi-task learning},
  author={Zhang, Yu and Yang, Qiang},
  journal={IEEE Transactions on Knowledge and Data Engineering},
  volume={34},
  number={12},
  pages={5586--5609},
  year={2021},
  publisher={IEEE}
}

@article{crawshaw2020multi,
  title={Multi-task learning with deep neural networks: A survey},
  author={Crawshaw, Michael},
  journal={arXiv preprint arXiv:2009.09796},
  year={2020}
}

@article{huh2016makes,
  title={What makes ImageNet good for transfer learning?},
  author={Huh, Minyoung and Agrawal, Pulkit and Efros, Alexei A},
  journal={arXiv preprint arXiv:1608.08614},
  year={2016}
}

@inproceedings{lee2022surgical,
  title={Surgical Fine-Tuning Improves Adaptation to Distribution Shifts},
  author={Lee, Yoonho and Chen, Annie S and Tajwar, Fahim and Kumar, Ananya and Yao, Huaxiu and Liang, Percy and Finn, Chelsea},
  booktitle={NeurIPS 2022 Workshop on Distribution Shifts: Connecting Methods and Applications}
}

@inproceedings{kumarfine,
  title={Fine-Tuning can Distort Pretrained Features and Underperform Out-of-Distribution},
  author={Kumar, Ananya and Raghunathan, Aditi and Jones, Robbie Matthew and Ma, Tengyu and Liang, Percy},
  booktitle={International Conference on Learning Representations}
}

@inproceedings{howard2018universal,
  title={Universal Language Model Fine-tuning for Text Classification},
  author={Howard, Jeremy and Ruder, Sebastian},
  booktitle={Proceedings of the 56th Annual Meeting of the Association for Computational Linguistics (Volume 1: Long Papers)},
  pages={328--339},
  year={2018}
}

@inproceedings{santoro2016meta,
  title={Meta-learning with memory-augmented neural networks},
  author={Santoro, Adam and Bartunov, Sergey and Botvinick, Matthew and Wierstra, Daan and Lillicrap, Timothy},
  booktitle={International conference on machine learning},
  pages={1842--1850},
  year={2016},
  organization={PMLR}
}

@inproceedings{mishra2017simple,
  title={A simple neural attentive meta-learner},
  author={Mishra, Nikhil and Rohaninejad, Mostafa and Chen, Xi and Abbeel, Pieter},
  booktitle={International Conference on Learning Representations},
  year={2018},
}

@inproceedings{munkhdalai2017meta,
  title={Meta networks},
  author={Munkhdalai, Tsendsuren and Yu, Hong},
  booktitle={International conference on machine learning},
  pages={2554--2563},
  year={2017},
  organization={PMLR}
}

@inproceedings{garnelo2018conditional,
  title={Conditional neural processes},
  author={Garnelo, Marta and Rosenbaum, Dan and Maddison, Christopher and Ramalho, Tiago and Saxton, David and Shanahan, Murray and Teh, Yee Whye and Rezende, Danilo and Eslami, SM Ali},
  booktitle={International conference on machine learning},
  pages={1704--1713},
  year={2018},
  organization={PMLR}
}

@inproceedings{finn2017model,
  title={Model-agnostic meta-learning for fast adaptation of deep networks},
  author={Finn, Chelsea and Abbeel, Pieter and Levine, Sergey},
  booktitle={International conference on machine learning},
  pages={1126--1135},
  year={2017},
  organization={PMLR}
}

@inproceedings{ravi2017optimization,
  title={Optimization as a model for few-shot learning},
  author={Ravi, Sachin and Larochelle, Hugo},
  booktitle={International conference on learning representations},
  year={2017}
}

@article{finn2017meta,
  title={Meta-learning and universality: Deep representations and gradient descent can approximate any learning algorithm},
  author={Finn, Chelsea and Levine, Sergey},
  booktitle={International Conference on Learning Representations},
  year={2018},
}

@article{li2017meta,
  title={Meta-sgd: Learning to learn quickly for few-shot learning},
  author={Li, Zhenguo and Zhou, Fengwei and Chen, Fei and Li, Hang},
  journal={arXiv preprint arXiv:1707.09835},
  year={2017}
}

@inproceedings{behl2019alpha,
  title={Alpha maml: Adaptive model-agnostic meta-learning},
  author={Behl, Harkirat Singh and Baydin, At{\i}l{\i}m G{\"u}ne{\c{s}} and Torr, Philip HS},
  booktitle={6th ICML Workshop on Automated Machine Learning, Thirty-Sixth International Conference on Machine Learning (ICML)},
  year={2019}
}

@article{zhou2018deep,
  title={Deep meta-learning: Learning to learn in the concept space},
  author={Zhou, Fengwei and Wu, Bin and Li, Zhenguo},
  journal={arXiv preprint arXiv:1802.03596},
  year={2018}
}

@inproceedings{zintgraf2019fast,
  title={Fast context adaptation via meta-learning},
  author={Zintgraf, Luisa and Shiarli, Kyriacos and Kurin, Vitaly and Hofmann, Katja and Whiteson, Shimon},
  booktitle={International Conference on Machine Learning},
  pages={7693--7702},
  year={2019},
  organization={PMLR}
}

@inproceedings{antoniou2018train,
  title={How to train your MAML},
  author={Antoniou, Antreas and Edwards, Harrison and Storkey, Amos},
  booktitle={International Conference on Learning Representations},
  year={2018}
}

@article{nichol2018first,
  title={On first-order meta-learning algorithms},
  author={Nichol, Alex and Achiam, Joshua and Schulman, John},
  journal={arXiv preprint arXiv:1803.02999},
  year={2018}
}

@inproceedings{bertinetto2018meta,
  title={Meta-learning with differentiable closed-form solvers},
  author={Bertinetto, Luca and Henriques, Joao F and Torr, Philip HS and Vedaldi, Andrea},
  booktitle={International Conference on Learning Representations (ICLR), 2019},
  year={2019}
}

@inproceedings{lee2019meta,
  title={Meta-learning with differentiable convex optimization},
  author={Lee, Kwonjoon and Maji, Subhransu and Ravichandran, Avinash and Soatto, Stefano},
  booktitle={Proceedings of the IEEE/CVF conference on computer vision and pattern recognition},
  pages={10657--10665},
  year={2019}
}

@article{rajeswaran2019meta,
  title={Meta-learning with implicit gradients},
  author={Rajeswaran, Aravind and Finn, Chelsea and Kakade, Sham M and Levine, Sergey},
  journal={Advances in neural information processing systems},
  volume={32},
  year={2019}
}

@inproceedings{zhang2018unreasonable,
  title={The unreasonable effectiveness of deep features as a perceptual metric},
  author={Zhang, Richard and Isola, Phillip and Efros, Alexei A and Shechtman, Eli and Wang, Oliver},
  booktitle={Proceedings of the IEEE conference on computer vision and pattern recognition},
  pages={586--595},
  year={2018}
}

@inproceedings{koch2015siamese,
  title={Siamese neural networks for one-shot image recognition},
  author={Koch, Gregory and Zemel, Richard and Salakhutdinov, Ruslan and others},
  booktitle={ICML deep learning workshop},
  volume={2},
  number={1},
  year={2015},
  organization={Lille}
}

@article{vinyals2016matching,
  title={Matching networks for one shot learning},
  author={Vinyals, Oriol and Blundell, Charles and Lillicrap, Timothy and Wierstra, Daan and others},
  journal={Advances in neural information processing systems},
  volume={29},
  year={2016}
}

@article{laenen2021episodes,
  title={On episodes, prototypical networks, and few-shot learning},
  author={Laenen, Steinar and Bertinetto, Luca},
  journal={Advances in Neural Information Processing Systems},
  volume={34},
  pages={24581--24592},
  year={2021}
}

@inproceedings{sung2018learning,
  title={Learning to compare: Relation network for few-shot learning},
  author={Sung, Flood and Yang, Yongxin and Zhang, Li and Xiang, Tao and Torr, Philip HS and Hospedales, Timothy M},
  booktitle={Proceedings of the IEEE conference on computer vision and pattern recognition},
  pages={1199--1208},
  year={2018}
}

@inproceedings{garcia2017few,
  title={Few-shot learning with graph neural networks},
  author={Garcia, Victor and Bruna, Joan},
  booktitle={International Conference on Learning Representations},
  year={2018},
}

@inproceedings{allen2019infinite,
  title={Infinite mixture prototypes for few-shot learning},
  author={Allen, Kelsey and Shelhamer, Evan and Shin, Hanul and Tenenbaum, Joshua},
  booktitle={International Conference on Machine Learning},
  pages={232--241},
  year={2019},
  organization={PMLR}
}

@inproceedings{jiang2019learning,
  title={Learning to learn with conditional class dependencies},
  author={Jiang, Xiang and Havaei, Mohammad and Varno, Farshid and Chartrand, Gabriel and Chapados, Nicolas and Matwin, Stan},
  booktitle={International conference on learning representations},
  year={2019}
}

@article{rusu2018meta,
  title={Meta-learning with latent embedding optimization},
  author={Rusu, Andrei A and Rao, Dushyant and Sygnowski, Jakub and Vinyals, Oriol and Pascanu, Razvan and Osindero, Simon and Hadsell, Raia},
  booktitle={International Conference on Learning Representations},
  year={2019},
}

@article{triantafillou2019meta,
  title={Meta-dataset: A dataset of datasets for learning to learn from few examples},
  author={Triantafillou, Eleni and Zhu, Tyler and Dumoulin, Vincent and Lamblin, Pascal and Evci, Utku and Xu, Kelvin and Goroshin, Ross and Gelada, Carles and Swersky, Kevin and Manzagol, Pierre-Antoine and others},
  booktitle={International Conference on Learning Representations},
  year={2020},
}

@article{oord2018representation,
  title={Representation learning with contrastive predictive coding},
  author={Oord, Aaron van den and Li, Yazhe and Vinyals, Oriol},
  journal={arXiv preprint arXiv:1807.03748},
  year={2018}
}

@inproceedings{chen2020simple,
  title={A simple framework for contrastive learning of visual representations},
  author={Chen, Ting and Kornblith, Simon and Norouzi, Mohammad and Hinton, Geoffrey},
  booktitle={International conference on machine learning},
  pages={1597--1607},
  year={2020},
  organization={PMLR}
}

@inproceedings{he2020momentum,
  title={Momentum contrast for unsupervised visual representation learning},
  author={He, Kaiming and Fan, Haoqi and Wu, Yuxin and Xie, Saining and Girshick, Ross},
  booktitle={Proceedings of the IEEE/CVF conference on computer vision and pattern recognition},
  pages={9729--9738},
  year={2020}
}

@article{grill2020bootstrap,
  title={Bootstrap your own latent-a new approach to self-supervised learning},
  author={Grill, Jean-Bastien and Strub, Florian and Altch{\'e}, Florent and Tallec, Corentin and Richemond, Pierre and Buchatskaya, Elena and Doersch, Carl and Avila Pires, Bernardo and Guo, Zhaohan and Gheshlaghi Azar, Mohammad and others},
  journal={Advances in neural information processing systems},
  volume={33},
  pages={21271--21284},
  year={2020}
}

@inproceedings{tamkin2020viewmaker,
  title={Viewmaker networks: Learning views for unsupervised representation learning},
  author={Tamkin, Alex and Wu, Mike and Goodman, Noah},
  booktitle={International Conference on Learning Representations},
  year={2020},
}

@inproceedings{nair2022r3m,
  title={R3m: A universal visual representation for robot manipulation},
  author={Nair, Suraj and Rajeswaran, Aravind and Kumar, Vikash and Finn, Chelsea and Gupta, Abhinav},
  booktitle={Conference on Robot Learning},
  pages={892--909},
  year={2023},
  organization={PMLR}
}

@inproceedings{ni2021close,
  title={The close relationship between contrastive learning and meta-learning},
  author={Ni, Renkun and Shu, Manli and Souri, Hossein and Goldblum, Micah and Goldstein, Tom},
  booktitle={International Conference on Learning Representations},
  year={2021}
}

@inproceedings{hsu2018unsupervised,
  title={Unsupervised learning via meta-learning},
  author={Hsu, Kyle and Levine, Sergey and Finn, Chelsea},
  booktitle={International Conference on Learning Representations},
  year={2019},
}

@inproceedings{donahue2016adversarial,
  title={Adversarial feature learning},
  author={Donahue, Jeff and Kr{\"a}henb{\"u}hl, Philipp and Darrell, Trevor},
  booktitle={International Conference on Learning Representations},
  year={2017},
}

@article{donahue2019large,
  title={Large scale adversarial representation learning},
  author={Donahue, Jeff and Simonyan, Karen},
  journal={Advances in neural information processing systems},
  volume={32},
  year={2019}
}

@inproceedings{caron2018deep,
  title={Deep clustering for unsupervised learning of visual features},
  author={Caron, Mathilde and Bojanowski, Piotr and Joulin, Armand and Douze, Matthijs},
  booktitle={Proceedings of the European conference on computer vision (ECCV)},
  pages={132--149},
  year={2018}
}

@article{khodadadeh2019unsupervised,
  title={Unsupervised meta-learning for few-shot image classification},
  author={Khodadadeh, Siavash and Boloni, Ladislau and Shah, Mubarak},
  journal={Advances in neural information processing systems},
  volume={32},
  year={2019}
}

@inproceedings{cubuk2019autoaugment,
  title={Autoaugment: Learning augmentation strategies from data},
  author={Cubuk, Ekin D and Zoph, Barret and Mane, Dandelion and Vasudevan, Vijay and Le, Quoc V},
  booktitle={Proceedings of the IEEE/CVF conference on computer vision and pattern recognition},
  pages={113--123},
  year={2019}
}

@article{brown2020language,
  title={Language models are few-shot learners},
  author={Brown, Tom and Mann, Benjamin and Ryder, Nick and Subbiah, Melanie and Kaplan, Jared D and Dhariwal, Prafulla and Neelakantan, Arvind and Shyam, Pranav and Sastry, Girish and Askell, Amanda and others},
  journal={Advances in neural information processing systems},
  volume={33},
  pages={1877--1901},
  year={2020}
}

@inproceedings{bansal2020self,
  title={Self-supervised meta-learning for few-shot natural language classification tasks},
  author={Bansal, Trapit and Jha, Rishikesh and Munkhdalai, Tsendsuren and McCallum, Andrew},
  booktitle = "Proceedings of the 2020 Conference on Empirical Methods in Natural Language Processing (EMNLP)",
  year={2020}
}

@article{tzeng2014deep,
  title={Deep domain confusion: Maximizing for domain invariance},
  author={Tzeng, Eric and Hoffman, Judy and Zhang, Ning and Saenko, Kate and Darrell, Trevor},
  journal={arXiv preprint arXiv:1412.3474},
  year={2014}
}

@article{ganin2016domain,
  title={Domain-adversarial training of neural networks},
  author={Ganin, Yaroslav and Ustinova, Evgeniya and Ajakan, Hana and Germain, Pascal and Larochelle, Hugo and Laviolette, Fran{\c{c}}ois and Marchand, Mario and Lempitsky, Victor},
  journal={The journal of machine learning research},
  volume={17},
  number={1},
  pages={2096--2030},
  year={2016},
  publisher={JMLR. org}
}

@inproceedings{zhu2017unpaired,
  title={Unpaired image-to-image translation using cycle-consistent adversarial networks},
  author={Zhu, Jun-Yan and Park, Taesung and Isola, Phillip and Efros, Alexei A},
  booktitle={Proceedings of the IEEE international conference on computer vision},
  pages={2223--2232},
  year={2017}
}

@inproceedings{rao2020rl,
  title={Rl-cyclegan: Reinforcement learning aware simulation-to-real},
  author={Rao, Kanishka and Harris, Chris and Irpan, Alex and Levine, Sergey and Ibarz, Julian and Khansari, Mohi},
  booktitle={Proceedings of the IEEE/CVF Conference on Computer Vision and Pattern Recognition},
  pages={11157--11166},
  year={2020}
}

@article{smith2019avid,
  title={Avid: Learning multi-stage tasks via pixel-level translation of human videos},
  author={Smith, Laura and Dhawan, Nikita and Zhang, Marvin and Abbeel, Pieter and Levine, Sergey},
  journal={arXiv preprint arXiv:1912.04443},
  year={2019}
}

@inproceedings{hoffman2018cycada,
  title={Cycada: Cycle-consistent adversarial domain adaptation},
  author={Hoffman, Judy and Tzeng, Eric and Park, Taesung and Zhu, Jun-Yan and Isola, Phillip and Saenko, Kate and Efros, Alexei and Darrell, Trevor},
  booktitle={International conference on machine learning},
  pages={1989--1998},
  year={2018},
  organization={Pmlr}
}

@article{sicilia2023domain,
  title={Domain adversarial neural networks for domain generalization: When it works and how to improve},
  author={Sicilia, Anthony and Zhao, Xingchen and Hwang, Seong Jae},
  journal={Machine Learning},
  pages={1--37},
  year={2023},
  publisher={Springer}
}

@inproceedings{sun2016deep,
  title={Deep coral: Correlation alignment for deep domain adaptation},
  author={Sun, Baochen and Saenko, Kate},
  booktitle={Computer Vision--ECCV 2016 Workshops: Amsterdam, The Netherlands, October 8-10 and 15-16, 2016, Proceedings, Part III 14},
  pages={443--450},
  year={2016},
  organization={Springer}
}

@inproceedings{zhang2017mixup,
  title={mixup: Beyond empirical risk minimization},
  author={Zhang, Hongyi and Cisse, Moustapha and Dauphin, Yann N and Lopez-Paz, David},
  booktitle={International Conference on Learning Representations},
  year={2018}
}

@inproceedings{yao2022improving,
  title={Improving out-of-distribution robustness via selective augmentation},
  author={Yao, Huaxiu and Wang, Yu and Li, Sai and Zhang, Linjun and Liang, Weixin and Zou, James and Finn, Chelsea},
  booktitle={International Conference on Machine Learning},
  pages={25407--25437},
  year={2022},
  organization={PMLR}
}

@inproceedings{verma2019manifold,
  title={Manifold mixup: Better representations by interpolating hidden states},
  author={Verma, Vikas and Lamb, Alex and Beckham, Christopher and Najafi, Amir and Mitliagkas, Ioannis and Lopez-Paz, David and Bengio, Yoshua},
  booktitle={International conference on machine learning},
  pages={6438--6447},
  year={2019},
  organization={PMLR}
}

@article{lopez2017gradient,
  title={Gradient episodic memory for continual learning},
  author={Lopez-Paz, David and Ranzato, Marc'Aurelio},
  journal={Advances in neural information processing systems},
  volume={30},
  year={2017}
}

@article{javed2019meta,
  title={Meta-learning representations for continual learning},
  author={Javed, Khurram and White, Martha},
  journal={Advances in neural information processing systems},
  volume={32},
  year={2019}
}

@inproceedings{beaulieu2020learning,
  title={Learning to continually learn},
  author={Beaulieu, Shawn and Frati, Lapo and Miconi, Thomas and Lehman, Joel and Stanley, Kenneth O and Clune, Jeff and Cheney, Nick},
  booktitle={24th European Conference on Artificial Intelligence},
  year={2020}
}

@inproceedings{prabhu2019few,
  title={Few-shot learning for dermatological disease diagnosis},
  author={Prabhu, Viraj and Kannan, Anitha and Ravuri, Murali and Chaplain, Manish and Sontag, David and Amatriain, Xavier},
  booktitle={Machine Learning for Healthcare Conference},
  pages={532--552},
  year={2019},
  organization={PMLR}
}

@inproceedings{liang2021cross,
  title={Cross-modal generalization: Learning in low resource modalities via meta-alignment},
  author={Liang, Paul Pu and Wu, Peter and Ziyin, Liu and Morency, Louis-Philippe and Salakhutdinov, Ruslan},
  booktitle={Proceedings of the 29th ACM International Conference on Multimedia},
  pages={2680--2689},
  year={2021}
}

@article{alayrac2022flamingo,
  title={Flamingo: a visual language model for few-shot learning},
  author={Alayrac, Jean-Baptiste and Donahue, Jeff and Luc, Pauline and Miech, Antoine and Barr, Iain and Hasson, Yana and Lenc, Karel and Mensch, Arthur and Millican, Katherine and Reynolds, Malcolm and others},
  journal={Advances in Neural Information Processing Systems},
  volume={35},
  pages={23716--23736},
  year={2022}
}

@article{reed2022generalist,
  title={A generalist agent},
  author={Reed, Scott and Zolna, Konrad and Parisotto, Emilio and Colmenarejo, Sergio Gomez and Novikov, Alexander and Barth-Maron, Gabriel and Gimenez, Mai and Sulsky, Yury and Kay, Jackie and Springenberg, Jost Tobias and others},
  journal={Transactions on Machine Learning Research},
  year={2022}
}

@article{ullah2022meta,
  title={Meta-album: Multi-domain meta-dataset for few-shot image classification},
  author={Ullah, Ihsan and Carri{\'o}n-Ojeda, Dustin and Escalera, Sergio and Guyon, Isabelle and Huisman, Mike and Mohr, Felix and van Rijn, Jan N and Sun, Haozhe and Vanschoren, Joaquin and Vu, Phan Anh},
  journal={Advances in Neural Information Processing Systems},
  volume={35},
  pages={3232--3247},
  year={2022}
}

@inproceedings{yu2020meta,
  title={Meta-world: A benchmark and evaluation for multi-task and meta reinforcement learning},
  author={Yu, Tianhe and Quillen, Deirdre and He, Zhanpeng and Julian, Ryan and Hausman, Karol and Finn, Chelsea and Levine, Sergey},
  booktitle={Conference on robot learning},
  pages={1094--1100},
  year={2020},
  organization={PMLR}
}

@article{zhai2019visual,
  title={The visual task adaptation benchmark},
  author={Zhai, Xiaohua and Puigcerver, Joan and Kolesnikov, Alexander and Ruyssen, Pierre and Riquelme, Carlos and Lucic, Mario and Djolonga, Josip and Pinto, Andre Susano and Neumann, Maxim and Dosovitskiy, Alexey and others},
  year={2019}
}

@inproceedings{zamir2018taskonomy,
  title={Taskonomy: Disentangling task transfer learning},
  author={Zamir, Amir R and Sax, Alexander and Shen, William and Guibas, Leonidas J and Malik, Jitendra and Savarese, Silvio},
  booktitle={Proceedings of the IEEE conference on computer vision and pattern recognition},
  pages={3712--3722},
  year={2018}
}

@inproceedings{li2021value,
  title={Value: A multi-task benchmark for video-and-language understanding evaluation},
  author={Li, Linjie and Lei, Jie and Gan, Zhe and Yu, Licheng and Chen, Yen-Chun and Pillai, Rohit and Cheng, Yu and Zhou, Luowei and Wang, Xin Eric and Wang, William Yang and others},
  booktitle={Thirty-fifth Conference on Neural Information Processing Systems Datasets and Benchmarks Track (Round 1)},
  year={2021},
}

@article{srivastava2022beyond,
  title={Beyond the imitation game: Quantifying and extrapolating the capabilities of language models},
  author={Srivastava, Aarohi and Rastogi, Abhinav and Rao, Abhishek and Shoeb, Abu Awal Md and Abid, Abubakar and Fisch, Adam and Brown, Adam R and Santoro, Adam and Gupta, Aditya and Garriga-Alonso, Adri{\`a} and others},
  journal={Transactions on Machine Learning Research},
  issn={2835-8856},
  year={2023},
}

@inproceedings{russwurm2020meta,
  title={Meta-learning for few-shot land cover classification},
  author={Ru{\ss}wurm, Marc and Wang, Sherrie and Korner, Marco and Lobell, David},
  booktitle={Proceedings of the ieee/cvf conference on computer vision and pattern recognition workshops},
  pages={200--201},
  year={2020}
}

@article{wu2021prototransformer,
  title={ProtoTransformer: A meta-learning approach to providing student feedback},
  author={Wu, Mike and Goodman, Noah and Piech, Chris and Finn, Chelsea},
  journal={arXiv preprint arXiv:2107.14035},
  year={2021}
}

@article{yu2018one,
  title={One-shot imitation from observing humans via domain-adaptive meta-learning},
  author={Yu, Tianhe and Finn, Chelsea and Xie, Annie and Dasari, Sudeep and Zhang, Tianhao and Abbeel, Pieter and Levine, Sergey},
  journal={Robotics: Science and Systems XIV},
  year={2018},
  publisher={Robotics: Science and Systems Foundation}
}

@inproceedings{nguyen2020meta,
  title={Meta-learning gnn initializations for low-resource molecular property prediction},
  author={Nguyen, Cuong Q and Kreatsoulas, Constantine and Branson, Kim M},
  booktitle={4th Lifelong Machine Learning Workshop at ICML 2020},
  year={2020},
}

@inproceedings{gui2018few,
  title={Few-shot human motion prediction via meta-learning},
  author={Gui, Liang-Yan and Wang, Yu-Xiong and Ramanan, Deva and Moura, Jos{\'e} MF},
  booktitle={Proceedings of the European Conference on Computer Vision (ECCV)},
  pages={432--450},
  year={2018}
}

@inproceedings{devlin2018bert,
  title={Bert: Pre-training of deep bidirectional transformers for language understanding},
  author={Kenton, Jacob Devlin Ming-Wei Chang and Toutanova, Lee Kristina},
  booktitle={Proceedings of NAACL-HLT},
  pages        = {4171--4186},
  publisher    = {Association for Computational Linguistics},
  year         = {2019},
}

@article{metz2020tasks,
  title={Tasks, stability, architecture, and compute: Training more effective learned optimizers, and using them to train themselves},
  author={Metz, Luke and Maheswaranathan, Niru and Freeman, C Daniel and Poole, Ben and Sohl-Dickstein, Jascha},
  journal={arXiv preprint arXiv:2009.11243},
  year={2020}
}

@article{karmaker2021automl,
  title={Automl to date and beyond: Challenges and opportunities},
  author={Karmaker, Shubhra Kanti and Hassan, Md Mahadi and Smith, Micah J and Xu, Lei and Zhai, Chengxiang and Veeramachaneni, Kalyan},
  journal={ACM Computing Surveys (CSUR)},
  volume={54},
  number={8},
  pages={1--36},
  year={2021},
  publisher={ACM New York, NY}
}

@article{vuorio2019multimodal,
  title={Multimodal model-agnostic meta-learning via task-aware modulation},
  author={Vuorio, Risto and Sun, Shao-Hua and Hu, Hexiang and Lim, Joseph J},
  journal={Advances in neural information processing systems},
  volume={32},
  year={2019}
}

@inproceedings{yao2019hierarchically,
  title={Hierarchically structured meta-learning},
  author={Yao, Huaxiu and Wei, Ying and Huang, Junzhou and Li, Zhenhui},
  booktitle={International Conference on Machine Learning},
  pages={7045--7054},
  year={2019},
  organization={PMLR}
}

@inproceedings{jiang2022subspace,
  title={Subspace learning for effective meta-learning},
  author={Jiang, Weisen and Kwok, James and Zhang, Yu},
  booktitle={International Conference on Machine Learning},
  pages={10177--10194},
  year={2022},
  organization={PMLR}
}

@article{jerfel2019reconciling,
  title={Reconciling meta-learning and continual learning with online mixtures of tasks},
  author={Jerfel, Ghassen and Grant, Erin and Griffiths, Tom and Heller, Katherine A},
  journal={Advances in Neural Information Processing Systems},
  volume={32},
  year={2019}
}

@inproceedings{zhou2021task,
  title={Task similarity aware meta learning: Theory-inspired improvement on maml},
  author={Zhou, Pan and Zou, Yingtian and Yuan, Xiao-Tong and Feng, Jiashi and Xiong, Caiming and Hoi, Steven},
  booktitle={Uncertainty in Artificial Intelligence},
  pages={23--33},
  year={2021},
  organization={PMLR}
}

@inproceedings{li2019lgm,
  title={LGM-Net: Learning to generate matching networks for few-shot learning},
  author={Li, Huaiyu and Dong, Weiming and Mei, Xing and Ma, Chongyang and Huang, Feiyue and Hu, Bao-Gang},
  booktitle={International conference on machine learning},
  pages={3825--3834},
  year={2019},
  organization={PMLR}
}

@article{requeima2019fast,
  title={Fast and flexible multi-task classification using conditional neural adaptive processes},
  author={Requeima, James and Gordon, Jonathan and Bronskill, John and Nowozin, Sebastian and Turner, Richard E},
  journal={Advances in Neural Information Processing Systems},
  volume={32},
  year={2019}
}

@inproceedings{liu2020universal,
  title={A Universal Representation Transformer Layer for Few-Shot Image Classification},
  author={Liu, Lu and Hamilton, William L and Long, Guodong and Jiang, Jing and Larochelle, Hugo},
  booktitle={International Conference on Learning Representations},
  year={2020}
}

@article{denevi2020advantage,
  title={The advantage of conditional meta-learning for biased regularization and fine tuning},
  author={Denevi, Giulia and Pontil, Massimiliano and Ciliberto, Carlo},
  journal={Advances in Neural Information Processing Systems},
  volume={33},
  pages={964--974},
  year={2020}
}

@inproceedings{triantafillou2021learning,
  title={Learning a universal template for few-shot dataset generalization},
  author={Triantafillou, Eleni and Larochelle, Hugo and Zemel, Richard and Dumoulin, Vincent},
  booktitle={International Conference on Machine Learning},
  pages={10424--10433},
  year={2021},
  organization={PMLR}
}

@inproceedings{dvornik2020selecting,
  title={Selecting relevant features from a multi-domain representation for few-shot classification},
  author={Dvornik, Nikita and Schmid, Cordelia and Mairal, Julien},
  booktitle={Computer Vision--ECCV 2020: 16th European Conference, Glasgow, UK, August 23--28, 2020, Proceedings, Part X 16},
  pages={769--786},
  year={2020},
  organization={Springer}
}

@inproceedings{li2021universal,
  title={Universal representation learning from multiple domains for few-shot classification},
  author={Li, Wei-Hong and Liu, Xialei and Bilen, Hakan},
  booktitle={Proceedings of the IEEE/CVF International Conference on Computer Vision},
  pages={9526--9535},
  year={2021}
}

@inproceedings{lee2022domain,
  title={Domain-Agnostic Meta-Learning for Cross-Domain Few-Shot Classification},
  author={Lee, Wei-Yu and Wang, Jheng-Yu and Wang, Yu-Chiang Frank},
  booktitle={ICASSP 2022-2022 IEEE International Conference on Acoustics, Speech and Signal Processing (ICASSP)},
  pages={1715--1719},
  year={2022},
  organization={IEEE}
}

@inproceedings{yuan2022task,
  title={Task-level Self-supervision for Cross-domain Few-shot Learning},
  author={Yuan, Wang and Zhang, Zhizhong and Wang, Cong and Song, Haichuan and Xie, Yuan and Ma, Lizhuang},
  booktitle={Proceedings of the AAAI Conference on Artificial Intelligence},
  volume={36},
  number={3},
  pages={3215--3223},
  year={2022}
}

@inproceedings{li2021semi,
  title={Semi-supervised meta-learning for cross-domain few-shot intent classification},
  author={Li, Yue and Zhang, Jiong},
  booktitle={Proceedings of the 1st Workshop on Meta Learning and Its Applications to Natural Language Processing},
  pages={67--75},
  year={2021}
}

@article{sener2018multi,
  title={Multi-task learning as multi-objective optimization},
  author={Sener, Ozan and Koltun, Vladlen},
  journal={Advances in neural information processing systems},
  volume={31},
  year={2018}
}

@inproceedings{kendall2018multi,
  title={Multi-task learning using uncertainty to weigh losses for scene geometry and semantics},
  author={Kendall, Alex and Gal, Yarin and Cipolla, Roberto},
  booktitle={Proceedings of the IEEE conference on computer vision and pattern recognition},
  pages={7482--7491},
  year={2018}
}

@inproceedings{ruder2019latent,
  title={Latent multi-task architecture learning},
  author={Ruder, Sebastian and Bingel, Joachim and Augenstein, Isabelle and S{\o}gaard, Anders},
  booktitle={Proceedings of the AAAI Conference on Artificial Intelligence},
  volume={33},
  number={01},
  pages={4822--4829},
  year={2019}
}

@inproceedings{gao2019nddr,
  title={Nddr-cnn: Layerwise feature fusing in multi-task cnns by neural discriminative dimensionality reduction},
  author={Gao, Yuan and Ma, Jiayi and Zhao, Mingbo and Liu, Wei and Yuille, Alan L},
  booktitle={Proceedings of the IEEE/CVF conference on computer vision and pattern recognition},
  pages={3205--3214},
  year={2019}
}

@inproceedings{oh2020boil,
  title={Boil: Towards representation change for few-shot learning},
  author={Oh, Jaehoon and Yoo, Hyungjun and Kim, ChangHwan and Yun, Se-Young},
  booktitle={The International Conference on Learning Representations (ICLR)},
  year={2021},
}

@inproceedings{mcmahan2017communication,
  title={Communication-efficient learning of deep networks from decentralized data},
  author={McMahan, Brendan and Moore, Eider and Ramage, Daniel and Hampson, Seth and y Arcas, Blaise Aguera},
  booktitle={Artificial intelligence and statistics},
  pages={1273--1282},
  year={2017},
  organization={PMLR}
}

@article{li2020federated,
  title={Federated optimization in heterogeneous networks},
  author={Li, Tian and Sahu, Anit Kumar and Zaheer, Manzil and Sanjabi, Maziar and Talwalkar, Ameet and Smith, Virginia},
  journal={Proceedings of Machine learning and systems},
  volume={2},
  pages={429--450},
  year={2020}
}

@inproceedings{li2021model,
  title={Model-contrastive federated learning},
  author={Li, Qinbin and He, Bingsheng and Song, Dawn},
  booktitle={Proceedings of the IEEE/CVF Conference on Computer Vision and Pattern Recognition},
  pages={10713--10722},
  year={2021}
}

@inproceedings{karimireddy2020scaffold,
  title={Scaffold: Stochastic controlled averaging for federated learning},
  author={Karimireddy, Sai Praneeth and Kale, Satyen and Mohri, Mehryar and Reddi, Sashank and Stich, Sebastian and Suresh, Ananda Theertha},
  booktitle={International Conference on Machine Learning},
  pages={5132--5143},
  year={2020},
  organization={PMLR}
}

@article{hanzely2020federated,
  title={Federated learning of a mixture of global and local models},
  author={Hanzely, Filip and Richt{\'a}rik, Peter},
  journal={arXiv preprint arXiv:2002.05516},
  year={2020}
}

@article{t2020personalized,
  title={Personalized federated learning with moreau envelopes},
  author={T Dinh, Canh and Tran, Nguyen and Nguyen, Josh},
  journal={Advances in Neural Information Processing Systems},
  volume={33},
  pages={21394--21405},
  year={2020}
}

@inproceedings{li2021ditto,
  title={Ditto: Fair and robust federated learning through personalization},
  author={Li, Tian and Hu, Shengyuan and Beirami, Ahmad and Smith, Virginia},
  booktitle={International Conference on Machine Learning},
  pages={6357--6368},
  year={2021},
  organization={PMLR}
}

@article{ghosh2020efficient,
  title={An efficient framework for clustered federated learning},
  author={Ghosh, Avishek and Chung, Jichan and Yin, Dong and Ramchandran, Kannan},
  journal={Advances in Neural Information Processing Systems},
  volume={33},
  pages={19586--19597},
  year={2020}
}

@article{duan2021flexible,
  title={Flexible clustered federated learning for client-level data distribution shift},
  author={Duan, Moming and Liu, Duo and Ji, Xinyuan and Wu, Yu and Liang, Liang and Chen, Xianzhang and Tan, Yujuan and Ren, Ao},
  journal={IEEE Transactions on Parallel and Distributed Systems},
  volume={33},
  number={11},
  pages={2661--2674},
  year={2021},
  publisher={IEEE}
}

@article{sattler2020clustered,
  title={Clustered federated learning: Model-agnostic distributed multitask optimization under privacy constraints},
  author={Sattler, Felix and M{\"u}ller, Klaus-Robert and Samek, Wojciech},
  journal={IEEE transactions on neural networks and learning systems},
  volume={32},
  number={8},
  pages={3710--3722},
  year={2020},
  publisher={IEEE}
}

@article{fallah2020personalized,
  title={Personalized federated learning with theoretical guarantees: A model-agnostic meta-learning approach},
  author={Fallah, Alireza and Mokhtari, Aryan and Ozdaglar, Asuman},
  journal={Advances in Neural Information Processing Systems},
  volume={33},
  pages={3557--3568},
  year={2020}
}

@inproceedings{fedpac2023,
  title={Personalized Federated Learning with Feature Alignment and Classifier Collaboration},
  author={Xu, Jian and Tong, Xinyi and Shao-Lun, Huang},
  booktitle={International conference on learning representations},
  year={2023}
}

@article{yang2023personalized,
  title={Personalized federated learning on non-IID data via group-based meta-learning},
  author={Yang, Lei and Huang, Jiaming and Lin, Wanyu and Cao, Jiannong},
  journal={ACM Transactions on Knowledge Discovery from Data},
  volume={17},
  number={4},
  pages={1--20},
  year={2023},
  publisher={ACM New York, NY}
}

@article{jiang2019improving,
  title={Improving federated learning personalization via model agnostic meta learning},
  author={Jiang, Yihan and Kone{\v{c}}n{\`y}, Jakub and Rush, Keith and Kannan, Sreeram},
  journal={arXiv preprint arXiv:1909.12488},
  year={2019}
}

@article{chen2018federated,
  title={Federated meta-learning with fast convergence and efficient communication},
  author={Chen, Fei and Luo, Mi and Dong, Zhenhua and Li, Zhenguo and He, Xiuqiang},
  journal={arXiv preprint arXiv:1802.07876},
  year={2018}
}

@article{khodak2019adaptive,
  title={Adaptive gradient-based meta-learning methods},
  author={Khodak, Mikhail and Balcan, Maria-Florina F and Talwalkar, Ameet S},
  journal={Advances in Neural Information Processing Systems},
  volume={32},
  year={2019}
}

@inproceedings{raghu2019rapid,
  title={Rapid learning or feature reuse? towards understanding the effectiveness of maml},
  author={Raghu, Aniruddh and Raghu, Maithra and Bengio, Samy and Vinyals, Oriol},
  booktitle={International conference on learning representations},
  year={2023}
}

@book{goodfellow2016deep,
  title={Deep learning},
  author={Goodfellow, Ian and Bengio, Yoshua and Courville, Aaron},
  year={2016},
  publisher={MIT press}
}

@article{hiller2022enforcing,
  title={On Enforcing Better Conditioned Meta-Learning for Rapid Few-Shot Adaptation},
  author={Hiller, Markus and Harandi, Mehrtash and Drummond, Tom},
  journal={Advances in Neural Information Processing Systems},
  year={2022}
}

@inproceedings{leeself,
  title={Self-Supervised Set Representation Learning for Unsupervised Meta-Learning},
  author={Lee, Dong Bok and Lee, Seanie and Kawaguchi, Kenji and Kim, Yunji and Bang, Jihwan and Ha, Jung-Woo and Hwang, Sung Ju},
  booktitle={International Conference on Learning Representations},
  year={2023},
}

@inproceedings{jang2023unsupervised,
  title={Unsupervised Meta-learning via Few-shot Pseudo-supervised Contrastive Learning},
  author={Jang, Huiwon and Lee, Hankook and Shin, Jinwoo},
  booktitle={International Conference on Learning Representations},
  year={2023}
}

@inproceedings{yang2022few,
  title={Few-shot classification with contrastive learning},
  author={Yang, Zhanyuan and Wang, Jinghua and Zhu, Yingying},
  booktitle={Computer Vision--ECCV 2022: 17th European Conference, Tel Aviv, Israel, October 23--27, 2022, Proceedings, Part XX},
  pages={293--309},
  year={2022},
  organization={Springer}
}

@inproceedings{khodadadeh2020unsupervised,
  title={Unsupervised meta-learning through latent-space interpolation in generative models},
  author={Khodadadeh, Siavash and Zehtabian, Sharare and Vahidian, Saeed and Wang, Weijia and Lin, Bill and B{\"o}l{\"o}ni, Ladislau},
  booktitle={International Conference on Learning Representations},
  year={2021}
}

@inproceedings{lee2021meta,
  title={Meta-gmvae: Mixture of gaussian vae for unsupervised meta-learning},
  author={Lee, Dong Bok and Min, Dongchan and Lee, Seanie and Hwang, Sung Ju},
  booktitle={International Conference on Learning Representations},
  year={2021}
}

@inproceedings{kong2021unsupervised,
  title={Unsupervised Meta-Learning via Latent Space Energy-based Model of Symbol Vector Coupling},
  author={Kong, Deqian and Pang, Bo and Wu, Ying Nian},
  booktitle={Fifth Workshop on Meta-Learning at the Conference on Neural Information Processing Systems},
  year={2021}
}

@inproceedings{kingma2013auto,
  title={Auto-encoding variational bayes},
  author={Kingma, Diederik P and Welling, Max},
  booktitle={International Conference on Learning Representations},
  year={2014}
}

@article{teh2003energy,
  title={Energy-based models for sparse overcomplete representations},
  author={Teh, Yee Whye and Welling, Max and Osindero, Simon and Hinton, Geoffrey E},
  journal={Journal of Machine Learning Research},
  volume={4},
  number={Dec},
  pages={1235--1260},
  year={2003}
}

@article{bansal2019learning,
  title={Learning to few-shot learn across diverse natural language classification tasks},
  author={Bansal, Trapit and Jha, Rishikesh and McCallum, Andrew},
  booktitle={Proceedings of the 28th International Conference on Computational Linguistics},
  pages={5108--5123},
  year={2020}
}

@inproceedings{wu2021one,
  title={One teacher is enough? pre-trained language model distillation from multiple teachers},
  author={Wu, Chuhan and Wu, Fangzhao and Huang, Yongfeng},
  booktitle={Findings of the Association for Computational Linguistics: ACL-IJCNLP 2021},
  pages={4408--4413},
  year={2021}
}

@article{zhao2018adversarial,
  title={Adversarial multiple source domain adaptation},
  author={Zhao, Han and Zhang, Shanghang and Wu, Guanhang and Moura, Jos{\'e} MF and Costeira, Joao P and Gordon, Geoffrey J},
  journal={Advances in neural information processing systems},
  volume={31},
  year={2018}
}

@inproceedings{nguyen2021unsupervised,
  title={Unsupervised multi-target domain adaptation through knowledge distillation},
  author={Nguyen-Meidine, Le Thanh and Belal, Atif and Kiran, Madhu and Dolz, Jose and Blais-Morin, Louis-Antoine and Granger, Eric},
  booktitle={Proceedings of the IEEE/CVF Winter Conference on Applications of Computer Vision},
  pages={1339--1347},
  year={2021}
}

@inproceedings{chen2019blending,
  title={Blending-target domain adaptation by adversarial meta-adaptation networks},
  author={Chen, Ziliang and Zhuang, Jingyu and Liang, Xiaodan and Lin, Liang},
  booktitle={Proceedings of the IEEE/CVF Conference on Computer Vision and Pattern Recognition},
  pages={2248--2257},
  year={2019}
}

@article{gholami2020unsupervised,
  title={Unsupervised multi-target domain adaptation: An information theoretic approach},
  author={Gholami, Behnam and Sahu, Pritish and Rudovic, Ognjen and Bousmalis, Konstantinos and Pavlovic, Vladimir},
  journal={IEEE Transactions on Image Processing},
  volume={29},
  pages={3993--4002},
  year={2020},
  publisher={IEEE}
}

@article{zhang2021adaptive,
  title={Adaptive risk minimization: Learning to adapt to domain shift},
  author={Zhang, Marvin and Marklund, Henrik and Dhawan, Nikita and Gupta, Abhishek and Levine, Sergey and Finn, Chelsea},
  journal={Advances in Neural Information Processing Systems},
  volume={34},
  pages={23664--23678},
  year={2021}
}

@inproceedings{yang2022fewda,
  title={Few-Shot Unsupervised Domain Adaptation via Meta Learning},
  author={Yang, Wanqi and Yang, Chengmei and Huang, Shengqi and Wang, Lei and Yang, Ming},
  booktitle={2022 IEEE International Conference on Multimedia and Expo (ICME)},
  pages={1--6},
  year={2022},
  organization={IEEE}
}

@article{feng2021similarity,
  title={Similarity-based meta-learning network with adversarial domain adaptation for cross-domain fault identification},
  author={Feng, Yong and Chen, Jinglong and Yang, Zhuozheng and Song, Xiaogang and Chang, Yuanhong and He, Shuilong and Xu, Enyong and Zhou, Zitong},
  journal={Knowledge-Based Systems},
  volume={217},
  pages={106829},
  year={2021},
  publisher={Elsevier}
}

@inproceedings{kang2018transferable,
  title={Transferable Meta Learning Across Domains.},
  author={Kang, Bingyi and Feng, Jiashi},
  booktitle={UAI},
  pages={177--187},
  year={2018}
}

@article{dou2019domain,
  title={Domain generalization via model-agnostic learning of semantic features},
  author={Dou, Qi and Coelho de Castro, Daniel and Kamnitsas, Konstantinos and Glocker, Ben},
  journal={Advances in Neural Information Processing Systems},
  volume={32},
  year={2019}
}

@article{balaji2018metareg,
  title={Metareg: Towards domain generalization using meta-regularization},
  author={Balaji, Yogesh and Sankaranarayanan, Swami and Chellappa, Rama},
  journal={Advances in neural information processing systems},
  volume={31},
  year={2018}
}

@inproceedings{li2018learning,
  title={Learning to generalize: Meta-learning for domain generalization},
  author={Li, Da and Yang, Yongxin and Song, Yi-Zhe and Hospedales, Timothy},
  booktitle={Proceedings of the AAAI conference on artificial intelligence},
  volume={32},
  number={1},
  year={2018}
}

@inproceedings{li2019episodic,
  title={Episodic training for domain generalization},
  author={Li, Da and Zhang, Jianshu and Yang, Yongxin and Liu, Cong and Song, Yi-Zhe and Hospedales, Timothy M},
  booktitle={Proceedings of the IEEE/CVF International Conference on Computer Vision},
  pages={1446--1455},
  year={2019}
}

@inproceedings{li2019feature,
  title={Feature-critic networks for heterogeneous domain generalization},
  author={Li, Yiying and Yang, Yongxin and Zhou, Wei and Hospedales, Timothy},
  booktitle={International Conference on Machine Learning},
  pages={3915--3924},
  year={2019},
  organization={PMLR}
}

@inproceedings{shu2021open,
  title={Open domain generalization with domain-augmented meta-learning},
  author={Shu, Yang and Cao, Zhangjie and Wang, Chenyu and Wang, Jianmin and Long, Mingsheng},
  booktitle={Proceedings of the IEEE/CVF Conference on Computer Vision and Pattern Recognition},
  pages={9624--9633},
  year={2021}
}

@article{chen2022discriminative,
  title={Discriminative adversarial domain generalization with meta-learning based cross-domain validation},
  author={Chen, Keyu and Zhuang, Di and Chang, J Morris},
  journal={Neurocomputing},
  volume={467},
  pages={418--426},
  year={2022},
  publisher={Elsevier}
}

@inproceedings{riemer2018learning,
  title={Learning to learn without forgetting by maximizing transfer and minimizing interference},
  author={Riemer, Matthew and Cases, Ignacio and Ajemian, Robert and Liu, Miao and Rish, Irina and Tu, Yuhai and Tesauro, Gerald},
  booktitle={International Conference on Learning Representations},
  year={2019},
}

@article{gupta2020look,
  title={Look-ahead meta learning for continual learning},
  author={Gupta, Gunshi and Yadav, Karmesh and Paull, Liam},
  journal={Advances in Neural Information Processing Systems},
  volume={33},
  pages={11588--11598},
  year={2020}
}

@inproceedings{chaudhry2018efficient,
  title={Efficient lifelong learning with a-gem},
  author={Chaudhry, Arslan and Ranzato, Marc'Aurelio and Rohrbach, Marcus and Elhoseiny, Mohamed},
  booktitle={International Conference on Learning Representations},
  year={2019},
}

@inproceedings{rebuffi2017icarl,
  title={icarl: Incremental classifier and representation learning},
  author={Rebuffi, Sylvestre-Alvise and Kolesnikov, Alexander and Sperl, Georg and Lampert, Christoph H},
  booktitle={Proceedings of the IEEE conference on Computer Vision and Pattern Recognition},
  pages={2001--2010},
  year={2017}
}

@inproceedings{aljundi2018selfless,
  title={Selfless sequential learning},
  author={Aljundi, Rahaf and Rohrbach, Marcus and Tuytelaars, Tinne},
  booktitle={International Conference on Learning Representations},
  year={2019}
}

@article{kirkpatrick2017overcoming,
  title={Overcoming catastrophic forgetting in neural networks},
  author={Kirkpatrick, James and Pascanu, Razvan and Rabinowitz, Neil and Veness, Joel and Desjardins, Guillaume and Rusu, Andrei A and Milan, Kieran and Quan, John and Ramalho, Tiago and Grabska-Barwinska, Agnieszka and others},
  journal={Proceedings of the national academy of sciences},
  volume={114},
  number={13},
  pages={3521--3526},
  year={2017},
  publisher={National Acad Sciences}
}

@inproceedings{serra2018overcoming,
  title={Overcoming catastrophic forgetting with hard attention to the task},
  author={Serra, Joan and Suris, Didac and Miron, Marius and Karatzoglou, Alexandros},
  booktitle={International Conference on Machine Learning},
  pages={4548--4557},
  year={2018},
  organization={PMLR}
}

@inproceedings{li2019learn,
  title={Learn to grow: A continual structure learning framework for overcoming catastrophic forgetting},
  author={Li, Xilai and Zhou, Yingbo and Wu, Tianfu and Socher, Richard and Xiong, Caiming},
  booktitle={International Conference on Machine Learning},
  pages={3925--3934},
  year={2019},
  organization={PMLR}
}

@inproceedings{pham2021contextual,
  title={Contextual transformation networks for online continual learning},
  author={Pham, Quang and Liu, Chenghao and Sahoo, Doyen and Steven, HOI},
  booktitle={International Conference on Learning Representations},
  year={2021}
}

@article{rusu2016progressive,
  title={Progressive neural networks},
  author={Rusu, Andrei A and Rabinowitz, Neil C and Desjardins, Guillaume and Soyer, Hubert and Kirkpatrick, James and Kavukcuoglu, Koray and Pascanu, Razvan and Hadsell, Raia},
  journal={arXiv preprint arXiv:1606.04671},
  year={2016}
}

@misc{vinalys,
  author =       "Oriol Vinyals",
  year =         "2017",
  title =        "Talk: Model vs Optimization Meta Learning",
  url =          "https://evolution.ml/pdf/vinyals.pdf",
  month =        dec,
  lastaccessed = "June 6, 2023",
  organization = {Neural Information Processing Systems (NIPS’17)}
}

@inbook{vanschoren2018meta,
  title={Meta-learning: A survey},
  author={Vanschoren, Joaquin},
  editor="Hutter, Frank and Kotthoff, Lars and Vanschoren, Joaquin",
  title="Meta-Learning",
  bookTitle="Automated Machine Learning: Methods, Systems, Challenges",
  year="2019",
  publisher="Springer International Publishing",
  pages="35--61",
}

@article{hospedales2021meta,
  title={Meta-learning in neural networks: A survey},
  author={Hospedales, Timothy and Antoniou, Antreas and Micaelli, Paul and Storkey, Amos},
  journal={IEEE transactions on pattern analysis and machine intelligence},
  volume={44},
  number={9},
  pages={5149--5169},
  year={2021},
  publisher={IEEE}
}

@article{vilalta2002perspective,
  title={A perspective view and survey of meta-learning},
  author={Vilalta, Ricardo and Drissi, Youssef},
  journal={Artificial intelligence review},
  volume={18},
  pages={77--95},
  year={2002},
  publisher={Springer}
}

@article{huisman2021survey,
  title={A survey of deep meta-learning},
  author={Huisman, Mike and Van Rijn, Jan N and Plaat, Aske},
  journal={Artificial Intelligence Review},
  volume={54},
  number={6},
  pages={4483--4541},
  year={2021},
  publisher={Springer}
}

@inproceedings{finn2017one,
  title={One-shot visual imitation learning via meta-learning},
  author={Finn, Chelsea and Yu, Tianhe and Zhang, Tianhao and Abbeel, Pieter and Levine, Sergey},
  booktitle={Conference on robot learning},
  pages={357--368},
  year={2017},
  organization={PMLR}
}

@inproceedings{madotto2019personalizing,
  title={Personalizing dialogue agents via meta-learning},
  author={Madotto, Andrea and Lin, Zhaojiang and Wu, Chien-Sheng and Fung, Pascale},
  booktitle={Proceedings of the 57th Annual Meeting of the Association for Computational Linguistics},
  pages={5454--5459},
  year={2019}
}

@inproceedings{qian2019domain,
  title={Domain Adaptive Dialog Generation via Meta Learning},
  author={Qian, Kun and Yu, Zhou},
  booktitle={Proceedings of the 57th Annual Meeting of the Association for Computational Linguistics},
  pages={2639--2649},
  year={2019}
}

@inproceedings{mi2019meta,
  title={Meta-learning for low-resource natural language generation in task-oriented dialogue systems},
  author={Mi, Fei and Huang, Minlie and Zhang, Jiyong and Faltings, Boi},
  booktitle={Proceedings of the 28th International Joint Conference on Artificial Intelligence},
  pages={3151--3157},
  year={2019}
}

@inproceedings{gu2020meta,
  title={Meta-learning for low-resource neural machine translation},
  author={Gu, Jiatao and Wang, Yong and Chen, Yun and Cho, Kyunghyun and Li, Victor OK},
  booktitle={2018 Conference on Empirical Methods in Natural Language Processing, EMNLP 2018},
  pages={3622--3631},
  year={2020},
  organization={Association for Computational Linguistics}
}

@inproceedings{geng2019induction,
  title={Induction Networks for Few-Shot Text Classification},
  author={Geng, Ruiying and Li, Binhua and Li, Yongbin and Zhu, Xiaodan and Jian, Ping and Sun, Jian},
  booktitle={Proceedings of the 2019 Conference on Empirical Methods in Natural Language Processing and the 9th International Joint Conference on Natural Language Processing (EMNLP-IJCNLP)},
  pages={3904--3913},
  year={2019}
}

@article{liang2023few,
  title={Few-shot aspect category sentiment analysis via meta-learning},
  author={Liang, Bin and Li, Xiang and Gui, Lin and Fu, Yonghao and He, Yulan and Yang, Min and Xu, Ruifeng},
  journal={ACM Transactions on Information Systems},
  volume={41},
  number={1},
  pages={1--31},
  year={2023},
  publisher={ACM New York, NY}
}

@article{snell2017prototypical,
  title={Prototypical networks for few-shot learning},
  author={Snell, Jake and Swersky, Kevin and Zemel, Richard},
  journal={Advances in neural information processing systems},
  volume={30},
  year={2017}
}

@article{touvron2023llama,
  title={Llama: Open and efficient foundation language models},
  author={Touvron, Hugo and Lavril, Thibaut and Izacard, Gautier and Martinet, Xavier and Lachaux, Marie-Anne and Lacroix, Timoth{\'e}e and Rozi{\`e}re, Baptiste and Goyal, Naman and Hambro, Eric and Azhar, Faisal and others},
  journal={arXiv preprint arXiv:2302.13971},
  year={2023}
}

@article{chowdhery2022palm,
  title={Palm: Scaling language modeling with pathways},
  author={Chowdhery, Aakanksha and Narang, Sharan and Devlin, Jacob and Bosma, Maarten and Mishra, Gaurav and Roberts, Adam and Barham, Paul and Chung, Hyung Won and Sutton, Charles and Gehrmann, Sebastian and others},
  journal={arXiv preprint arXiv:2204.02311},
  year={2022}
}

@article{OpenAI2023GPT4TR,
  title={GPT-4 Technical Report},
  author={OpenAI},
  journal={ArXiv},
  year={2023},
  volume={abs/2303.08774}
}

@article{phang2018sentence,
  title={Sentence encoders on stilts: Supplementary training on intermediate labeled-data tasks},
  author={Phang, Jason and F{\'e}vry, Thibault and Bowman, Samuel R},
  journal={arXiv preprint arXiv:1811.01088},
  year={2018}
}

@inproceedings{tian2020rethinking,
  title={Rethinking few-shot image classification: a good embedding is all you need?},
  author={Tian, Yonglong and Wang, Yue and Krishnan, Dilip and Tenenbaum, Joshua B and Isola, Phillip},
  booktitle={Computer Vision--ECCV 2020: 16th European Conference, Glasgow, UK, August 23--28, 2020, Proceedings, Part XIV 16},
  pages={266--282},
  year={2020},
  organization={Springer}
}

@inproceedings{Chen_2021_ICCV,
    author    = {Chen, Yinbo and Liu, Zhuang and Xu, Huijuan and Darrell, Trevor and Wang, Xiaolong},
    title     = {Meta-Baseline: Exploring Simple Meta-Learning for Few-Shot Learning},
    booktitle = {Proceedings of the IEEE/CVF International Conference on Computer Vision (ICCV)},
    year      = {2021},
    pages     = {9062-9071}
}

@inproceedings{guo2020broader,
  title={A broader study of cross-domain few-shot learning},
  author={Guo, Yunhui and Codella, Noel C and Karlinsky, Leonid and Codella, James V and Smith, John R and Saenko, Kate and Rosing, Tajana and Feris, Rogerio},
  booktitle={Computer Vision--ECCV 2020: 16th European Conference, Glasgow, UK, August 23--28, 2020, Proceedings, Part XXVII 16},
  pages={124--141},
  year={2020},
  organization={Springer}
}

@inproceedings{chen2019closer,
  title={A Closer Look at Few-shot Classification},
  author={Chen, Wei-Yu and Liu, Yen-Cheng and Kira, Zsolt and Wang, Yu-Chiang Frank and Huang, Jia-Bin},
  booktitle={International Conference on Learning Representations},
  year={2019}
}

@article{andrychowicz2016learning,
  title={Learning to learn by gradient descent by gradient descent},
  author={Andrychowicz, Marcin and Denil, Misha and Gomez, Sergio and Hoffman, Matthew W and Pfau, David and Schaul, Tom and Shillingford, Brendan and De Freitas, Nando},
  journal={Advances in neural information processing systems},
  volume={29},
  year={2016}
}

@inproceedings{
li2017learning,
title={Learning to Optimize},
author={Ke Li and Jitendra Malik},
booktitle={International Conference on Learning Representations},
year={2017},
}

@inproceedings{wichrowska2017learned,
  title={Learned optimizers that scale and generalize},
  author={Wichrowska, Olga and Maheswaranathan, Niru and Hoffman, Matthew W and Colmenarejo, Sergio Gomez and Denil, Misha and Freitas, Nando and Sohl-Dickstein, Jascha},
  booktitle={International Conference on Machine Learning},
  pages={3751--3760},
  year={2017},
  organization={PMLR}
}

@inproceedings{franceschi2018bilevel,
  title={Bilevel programming for hyperparameter optimization and meta-learning},
  author={Franceschi, Luca and Frasconi, Paolo and Salzo, Saverio and Grazzi, Riccardo and Pontil, Massimiliano},
  booktitle={International conference on machine learning},
  pages={1568--1577},
  year={2018},
  organization={PMLR}
}

@article{shaw2019meta,
  title={Meta architecture search},
  author={Shaw, Albert and Wei, Wei and Liu, Weiyang and Song, Le and Dai, Bo},
  journal={Advances in Neural Information Processing Systems},
  volume={32},
  year={2019}
}

@inproceedings{lian2019towards,
  title={Towards fast adaptation of neural architectures with meta learning},
  author={Lian, Dongze and Zheng, Yin and Xu, Yintao and Lu, Yanxiong and Lin, Leyu and Zhao, Peilin and Huang, Junzhou and Gao, Shenghua},
  booktitle={International Conference on Learning Representations},
  year={2019}
}

@article{wang2019hybrid,
  title={A hybrid approach with optimization-based and metric-based meta-learner for few-shot learning},
  author={Wang, Duo and Cheng, Yu and Yu, Mo and Guo, Xiaoxiao and Zhang, Tao},
  journal={Neurocomputing},
  volume={349},
  pages={202--211},
  year={2019},
  publisher={Elsevier}
}

@article{van2022three,
  title={Three types of incremental learning},
  author={van de Ven, Gido M and Tuytelaars, Tinne and Tolias, Andreas S},
  journal={Nature Machine Intelligence},
  volume={4},
  number={12},
  pages={1185--1197},
  year={2022},
  publisher={Nature Publishing Group UK London}
}

@article{bornschein2022nevis,
  title={NEVIS'22: A Stream of 100 Tasks Sampled from 30 Years of Computer Vision Research},
  author={Bornschein, Jorg and Galashov, Alexandre and Hemsley, Ross and Rannen-Triki, Amal and Chen, Yutian and Chaudhry, Arslan and He, Xu Owen and Douillard, Arthur and Caccia, Massimo and Feng, Qixuang and others},
  journal={arXiv preprint arXiv:2211.11747},
  year={2022}
}

@article{Cover1967,
  author={Cover, T. and Hart, P.},
  journal={IEEE Transactions on Information Theory}, 
  title={Nearest neighbor pattern classification}, 
  year={1967},
  volume={13},
  number={1},
  pages={21-27},
}

@article{garg2022can,
  title={What can transformers learn in-context? a case study of simple function classes},
  author={Garg, Shivam and Tsipras, Dimitris and Liang, Percy S and Valiant, Gregory},
  journal={Advances in Neural Information Processing Systems},
  volume={35},
  pages={30583--30598},
  year={2022}
}

@inproceedings{kirsch2022general,
  title={General-purpose in-context learning by meta-learning transformers},
  author={Kirsch, Louis and Harrison, James and Sohl-Dickstein, Jascha and Metz, Luke},
  booktitle={NeurIPS 2022 Workshop on Distribution Shifts: Connecting Methods and Applications},
  year={2022}
}

@inproceedings{akyurek2022learning,
  title={What learning algorithm is in-context learning? Investigations with linear models},
  author={Aky{\"u}rek, Ekin and Schuurmans, Dale and Andreas, Jacob and Ma, Tengyu and Zhou, Denny},
  booktitle={The Eleventh International Conference on Learning Representations},
  year={2022}
}

@article{vaswani2017attention,
  title={Attention is all you need},
  author={Vaswani, Ashish and Shazeer, Noam and Parmar, Niki and Uszkoreit, Jakob and Jones, Llion and Gomez, Aidan N and Kaiser, {\L}ukasz and Polosukhin, Illia},
  journal={Advances in neural information processing systems},
  volume={30},
  year={2017}
}

@incollection{ZOU20231,
title = {Chapter 1 - Meta-learning basics and background},
editor = {Lan Zou},
booktitle = {Meta-Learning},
publisher = {Academic Press},
pages = {1-22},
year = {2023},
author = {Lan Zou},
}

@article{kim2018bayesian,
  title={Bayesian model-agnostic meta-learning},
  author={Kim, Taesup and Yoon, Jaesik and Dia, Ousmane and Kim, Sungwoong and Bengio, Yoshua and Ahn, Sungjin},
  journal={Advances in neural information processing systems},
  year={2018}
}

@article{chen2020closer,
  title={A closer look at the training strategy for modern meta-learning},
  author={Chen, Jiaxin and Wu, Xiao-Ming and Li, Yanke and Li, Qimai and Zhan, Li-Ming and Chung, Fu-lai},
  journal={Advances in Neural Information Processing Systems},
  volume={33},
  pages={396--406},
  year={2020}
}

@inproceedings{lucas2020theoretical,
  title={Theoretical bounds on estimation error for meta-learning},
  author={Lucas, James and Ren, Mengye and KAMENI, Irene Raissa KAMENI and Pitassi, Toniann and Zemel, Richard},
  booktitle={International Conference on Learning Representations},
  year={2020}
}

@inproceedings{guan2022fast,
  title={Fast-rate PAC-Bayesian generalization bounds for meta-learning},
  author={Guan, Jiechao and Lu, Zhiwu},
  booktitle={International Conference on Machine Learning},
  pages={7930--7948},
  year={2022},
  organization={PMLR}
}

@article{beck2023survey,
  title={A survey of meta-reinforcement learning},
  author={Beck, Jacob and Vuorio, Risto and Liu, Evan Zheran and Xiong, Zheng and Zintgraf, Luisa and Finn, Chelsea and Whiteson, Shimon},
  journal={arXiv preprint arXiv:2301.08028},
  year={2023}
}

@inproceedings{vettoruzzo2023meta,
  title={Meta-Learning from Multimodal Task Distributions Using Multiple Sets of Meta-Parameters},
  author={Vettoruzzo, Anna and Bouguelia, Mohamed-Rafik and R{\"o}gnvaldsson, Thorsteinn},
  booktitle={2023 International Joint Conference on Neural Networks (IJCNN)},
  pages={1--8},
  year={2023},
  organization={IEEE}
}
\balance
\vskip -2\baselineskip plus -1fil
\begin{IEEEbiography}
[{\includegraphics[width=1in,height=1.25in,clip,keepaspectratio]{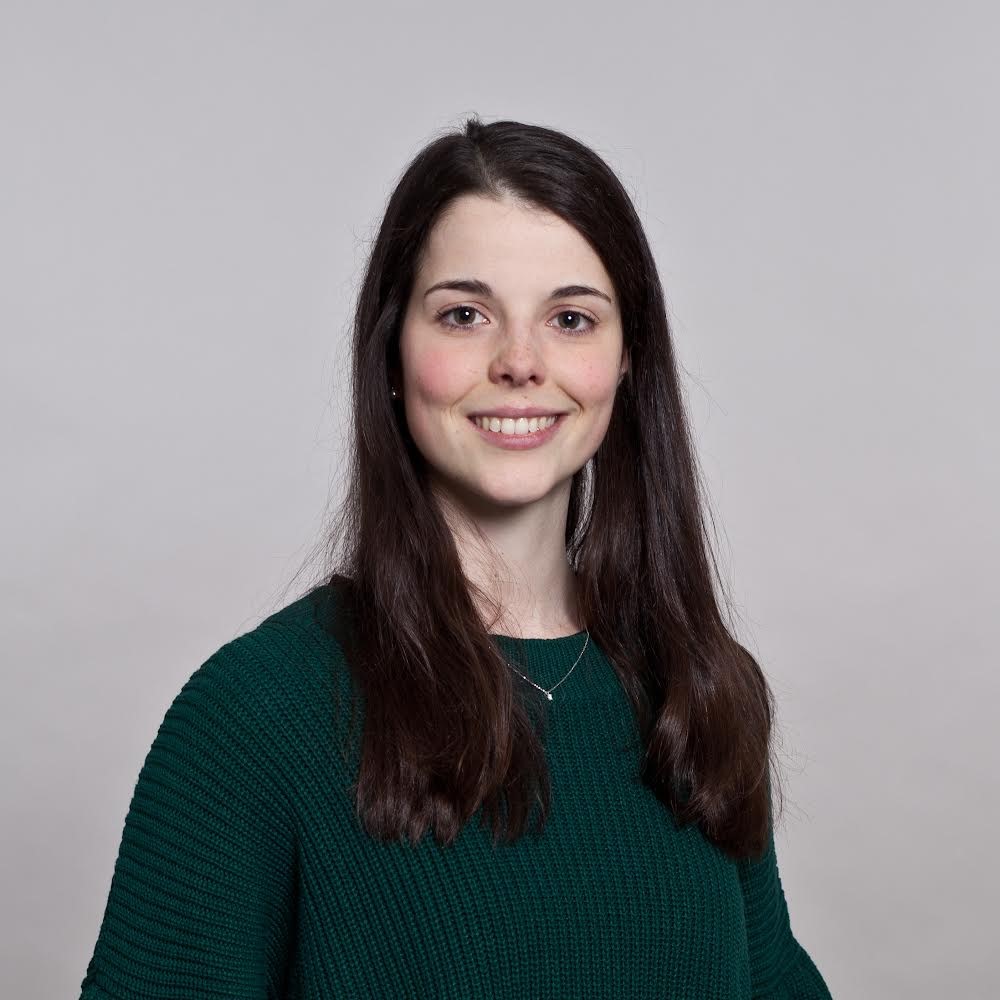}}]{Anna Vettoruzzo}
is a Ph.D. student at the Center of Applied and Intelligence System Research at Halmstad University (Sweden). She received her M.Sc. degree in ICT for Internet and Multimedia at the University of Padova in 2019 with focus on Machine Learning for Healthcare. Her research interests cover machine learning algorithms that enable machines with the ability of learning-to-learn. 
\end{IEEEbiography}
\vskip -2\baselineskip plus -1fil
\begin{IEEEbiography}
[{\includegraphics[width=1in,height=1.25in,clip,keepaspectratio]{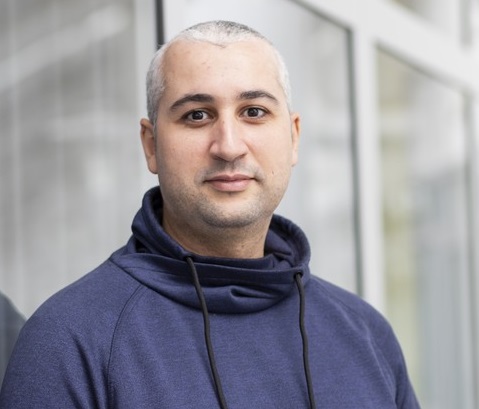}}]{Mohamed-Rafik Bouguelia}
is an Associate Professor and Docent in Machine Learning at Halmstad University, Sweden. He is also the program manager for the Applied AI program at the university. Previously, he conducted research at the University of Lorraine and the INRIA research center in France, earning his Ph.D. in Computer Science with a focus on Machine Learning. He holds an M.Sc. in Computer Science from USTHB University, Algeria. Bouguelia's current research interests encompass interactive machine learning, representation learning with deep neural networks, transfer learning and meta-learning.
\end{IEEEbiography}
\vskip -2\baselineskip plus -1fil
\begin{IEEEbiography}
[{\includegraphics[width=1in,height=1.25in,clip,keepaspectratio]{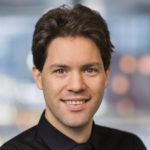}}]{Joaquin Vanschoren} is an Associate Professor of Machine Learning at Eindhoven University of Technology (TU/e). His research focuses on understanding and automating machine learning, meta-learning, and continual learning. He founded OpenML.org, the NeurIPS Datasets and Benchmark track, the DMLR journal, and the AutoML and meta-learning workshops at ICML and NeurIPS. He won the Dutch Data Prize and an Amazon Researcher Award. He was tutorial speaker at NeurIPS 2018 and AAAI 2021, and gave over 30 invited talks. He authored and co-authored over 150 scientific papers, as well as reference books on Automated Machine Learning and Meta-learning. He is a founding member of the European AI networks ELLIS and CLAIRE.
\end{IEEEbiography}
\vskip -2\baselineskip plus -1fil
\begin{IEEEbiography}
[{\includegraphics[width=1in,height=1.25in,clip,keepaspectratio]{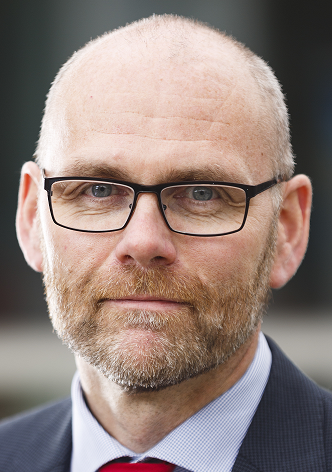}}]{Thorsteinn Rögnvaldsson}
is a senior member of IEEE and professor of computer science at Halmstad University, Sweden. From 2012, he started and directed the Center for Applied Intelligent Systems Research (CAISR) at Halmstad University. He has a PhD in theoretical physics from Lund University 1994 and did his post-doc at the Oregon Graduate Institute. His research interests are in autonomous knowledge creation, machine learning and self-organization. 
\end{IEEEbiography}
\vskip -2\baselineskip plus -1fil
\begin{IEEEbiography}
[{\includegraphics[width=1in,height=1.25in,clip,keepaspectratio]{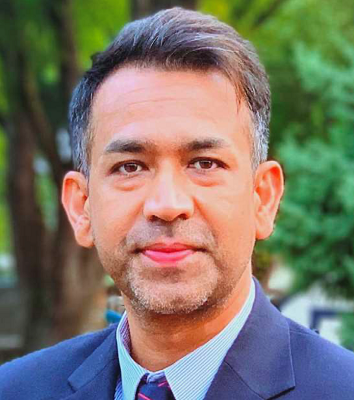}}] {KC Santosh} - a highly accomplished AI expert - is the chair of the Department of Computer at the University of South Dakota (USD), holding a Ph.D. in Computer Science - AI from INRIA Nancy Grand East Research Centre (France). Immediately after his postdoc at the LORIA reseach centre, Universit`e de Lorraine, he joined the National Institutes of Health as a research fellow. With funding of over \$1.3 million, including a \$1 million grant from DoD (2023) for AI/ML capacity building at USD, he has authored 10 books and published over 240 peer-reviewed research articles. He is also an editor of multiple prestigious journals such as IEEE Transactions on AI, Int. J of Machine Learning \& Cybernetics, and Int. J of Pattern Recognition \& AI. As founder of USD's AI programs, he significantly boosted graduate program enrollment by over 3,000\% in three years, establishing inter-disciplinary AI/Data Science programs with various departments. Prof. Santosh's leadership positions USD as an AI pioneer in South Dakota. Learn more: \url{https://kc-santosh.org/}.
\end{IEEEbiography}

\end{document}